\documentclass[10pt,journal,compsoc]{IEEEtran}

\ifCLASSINFOpdf
\else
\fi
\usepackage{amsmath}
\usepackage{url}
\usepackage{graphicx}
\graphicspath{{images/}}

\usepackage{xspace}
\usepackage{mathrsfs}
\usepackage{amssymb}
\usepackage{bm}
\usepackage{listings}
\usepackage{hyperref}
\usepackage{array,booktabs}
\usepackage{times}
\usepackage{epsfig}
\usepackage{multirow}
\usepackage{ulem}
\usepackage[dvipsnames]{xcolor}
\usepackage{boldline}
\usepackage{soul}
\usepackage{float}

\makeatletter
\newcommand\extralabel[2]{{\edef\@currentlabel{\@currentlabel#2}\label{#1}}}
\makeatother

\newcommand{\R}{\mathbb R}
\newcommand{\N}{\mathbb N}

\newcolumntype{C}[1]{>{\centering\let\newline\\\arraybackslash\hspace{0pt}}p{#1}}
\newcolumntype{L}[1]{>{\let\newline\\\arraybackslash\hspace{0pt}}p{#1}}

\newcommand{\methname}{ POMP-BE-PD\xspace}

\makeatletter
\DeclareRobustCommand\onedot{\futurelet\@let@token\@onedot}
\def\@onedot{\ifx\@let@token.\else.\null\fi\xspace}

\def\eg{\textit{e.g}\onedot} 
\def\ie{\textit{i.e}\onedot}

\def\wrt{w.r.t\onedot} 
\def\etal{\textit{et al}\onedot}
\makeatother

\begin{document}
%
\title{Unsupervised Active Visual Search with Monte Carlo planning under Uncertain Detections}
%
%
%

\author{Francesco~Taioli,
        Francesco~Giuliari,
        Yiming~Wang,
        Riccardo~Berra,
        Alberto~Castellini,
        Alessio~Del~Bue,~\IEEEmembership{Member,~IEEE,}
        Alessandro~Farinelli, 
        Marco~Cristani,~\IEEEmembership{Member,~IEEE,}
        and Francesco~Setti,~\IEEEmembership{Member,~IEEE}
\IEEEcompsocitemizethanks{\IEEEcompsocthanksitem F. Taioli, R. Berra, A. Castellini, A. Farinelli, M. Cristani and F. Setti are with the Department of Engineering for Innovation Medicine, University of Verona, Verona, Italy.
\IEEEcompsocthanksitem F. Giuliari is with University of Genoa (UniGe) and the Pattern Analysis and Computer Vision (PAVIS) research line, Istituto Italiano di Tecnologia (IIT), Genova, Italy.%
\IEEEcompsocthanksitem A. Del Bue is with the Pattern Analysis and Computer Vision (PAVIS) research line, Istituto Italiano di Tecnologia (IIT), Genova, Italy.%
\IEEEcompsocthanksitem Y. Wang is with Deep Visual Learning (DVL) unit, Fondazione Bruno Kessler (FBK), Trento, Italy.%
}


}

%
%

\markboth{}{}


%



\maketitle

\begin{abstract}
We propose a solution for Active Visual Search of objects in an environment, whose 2D floor map is the only known information. Our solution has three key features that make it more plausible and robust to detector failures compared to state-of-the-art methods: 
\textit{(i)} it~is unsupervised as it does not need any training sessions. \textit{(ii)}~During the exploration, a probability distribution on the 2D floor map is updated according to an intuitive mechanism, while an improved belief update increases the effectiveness of the agent’s exploration.
\textit{(iii)} We incorporate the awareness that an object detector may fail into the aforementioned probability modelling by exploiting the success statistics of a specific detector. Our solution is dubbed \methname (Pomcp-based Online Motion Planning with Belief by Exploration and Probabilistic Detection). It uses the current pose of an agent and an RGB-D observation to learn an optimal search policy, exploiting a POMDP solved by a Monte-Carlo planning approach. On the Active Vision Database benchmark, we increase the average success rate over all the environments by a significant 35$\%$ while decreasing the average path length by 4$\%$ with respect to competing methods. Thus, our results are state-of-the-art, even without using any training procedure.
\end{abstract}



%
\IEEEpeerreviewmaketitle

\section{Introduction}
\label{sec:intro}
\begin{figure}[!t]
    \centering
    \includegraphics[width=1\linewidth]{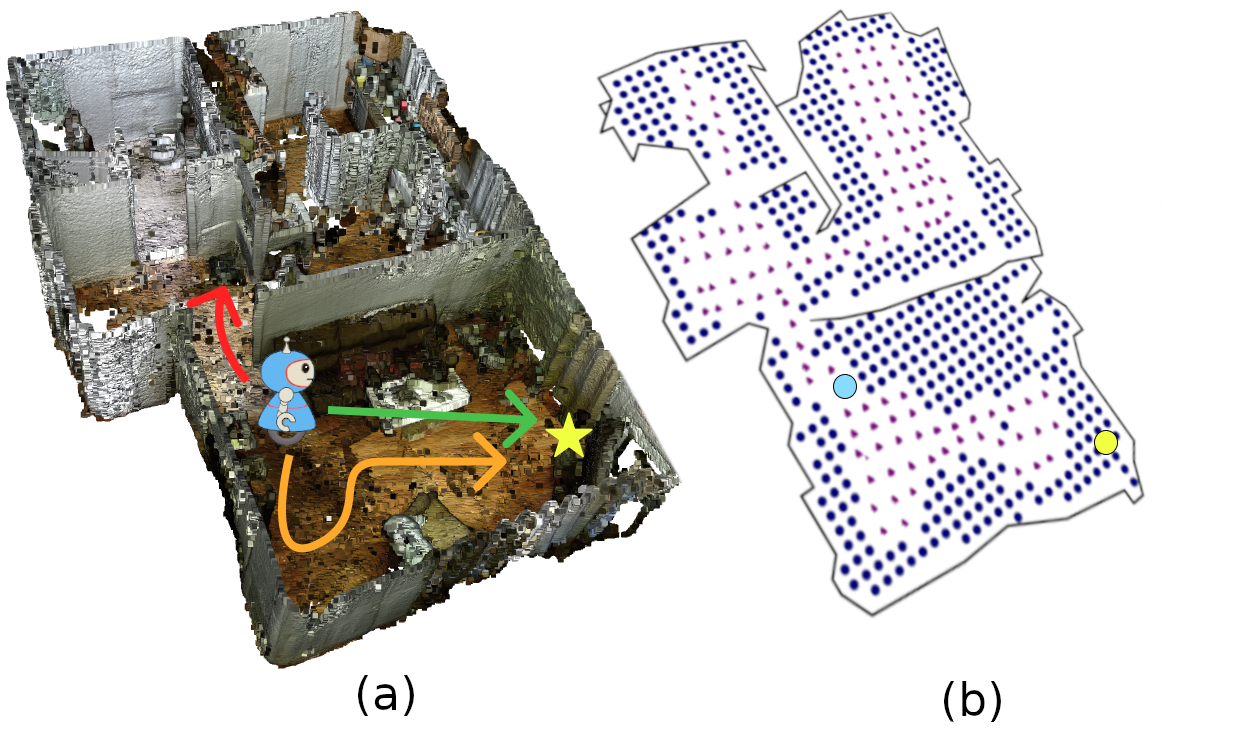}
    \caption{An agent is initialised in a known environment with the task of visually searching for a target object, \ie to localise the object and approach it. (a) 3D reconstruction of the environment; the agent has to navigate toward the target (yellow star) through the possible shortest path (highlighted in green) while avoiding longer trajectories (in orange) without missing entirely the target (in red). (b) Corresponding 2D grid map of the scene in our POMCP modelling: blue dots are the possible object locations, purple crosses are the possible robot poses.}
    \label{fig:intro_figure}
\end{figure}

\IEEEPARstart{A}{mong} the most interesting areas of robotics is the problem of Active Visual Search (AVS)~\cite{tsotsos1992relative}, in which an intelligent robotic agent must autonomously find an object located far from it, moving and exploring its surroundings through egocentric visual sensors. AVS applies in many different contexts, such as the domotic field~\cite{sjoo2012topological,giuliari2022spatial}, personal assistance~\cite{park2022zero}, search and rescue~\cite{rasouli2020attention,leslie2022robots}, and the very intriguing Mars exploration~\cite{pouya2021performing}.
%
%
In this paper we focus on the AVS problem in an indoor environment~\cite{ammirato2017dataset}, where the only available knowledge is its 2D map.
We propose a method for synthesising a motion planning policy that decides how to move an agent based on a perception module to visually detect and approach a specific object, \ie the target (see Fig.~\ref{fig:intro_figure}). 

AVS in real-world scenarios with egocentric camera views is a very challenging problem due to the unpredictable quality of the observations --\ie object in the far field, motion blur and low resolution--, partial views and occlusions due to scene clutters and generalisation to new environment.
This has an impact not only on the object detection but also on the planning policy.
To address this challenge, recent efforts are mostly based on deep Reinforcement Learning (RL), \eg deep recurrent Q-network (DRQN), fed with deep visual embedding~\cite{schmid2019iros,ye2019ral}. To train such DRQN models, a large amount of data is required, which are sequences of observations of various lengths, covering successful and unsuccessful search episodes from multiple real scenarios or simulated environments.

In this paper we take a different perspective and propose an online learning method for AVS. The basic idea is not to learn a complete policy by using a vast amount of training data, instead to use an advanced planning approach that can devise the best action based on the environment configuration which is built starting from the observations gathered in the environment.  

This fundamental shift in the methodology is carried out considering the Partially Observable Monte Carlo Planning (POMCP) method~\cite{Silver2010}. POMCP has been applied in benchmark problems, such as rocksample, battleship and \textit{pocman} (partially observable pacman)  with impressive results, however, its use for robotic applications is an open and challenging research problem.

In our previous work~\cite{pomp2020bmvc}, we introduced POMP, an online policy learning method, that uses as input the current pose of an agent and an RGB-D frame to plan the next move that brings the agent closer to the target object.
We modelled the problem as a Partially Observable Markov Decision Process solved by a Monte-Carlo planning approach, allowing us to explore the environment and search for the object at the same time.
The main benefit of this approach is that, differently from Reinforcement Learning approaches, POMP does not require a training phase, so being very agile in solving AVS in small and medium real scenarios.
Despite achieving results close to the state-of-the-art without using any training data, real object detectors are inaccurate with false positives and miss-detections. As a consequence, the agent could terminate the exploration in wrong locations, fooled by the detector, and thus decreasing the  overall success rate.

To overcome these problems, in this paper we propose \methname, an extension of POMP that poses the observation model in probabilistic terms, allowing us to better handle the false positives of a realistic object detector, as well as improving the effectiveness of the agent's explorations.

\begin{figure*}[t!]
  \centering
  \includegraphics[width=.7\textwidth]{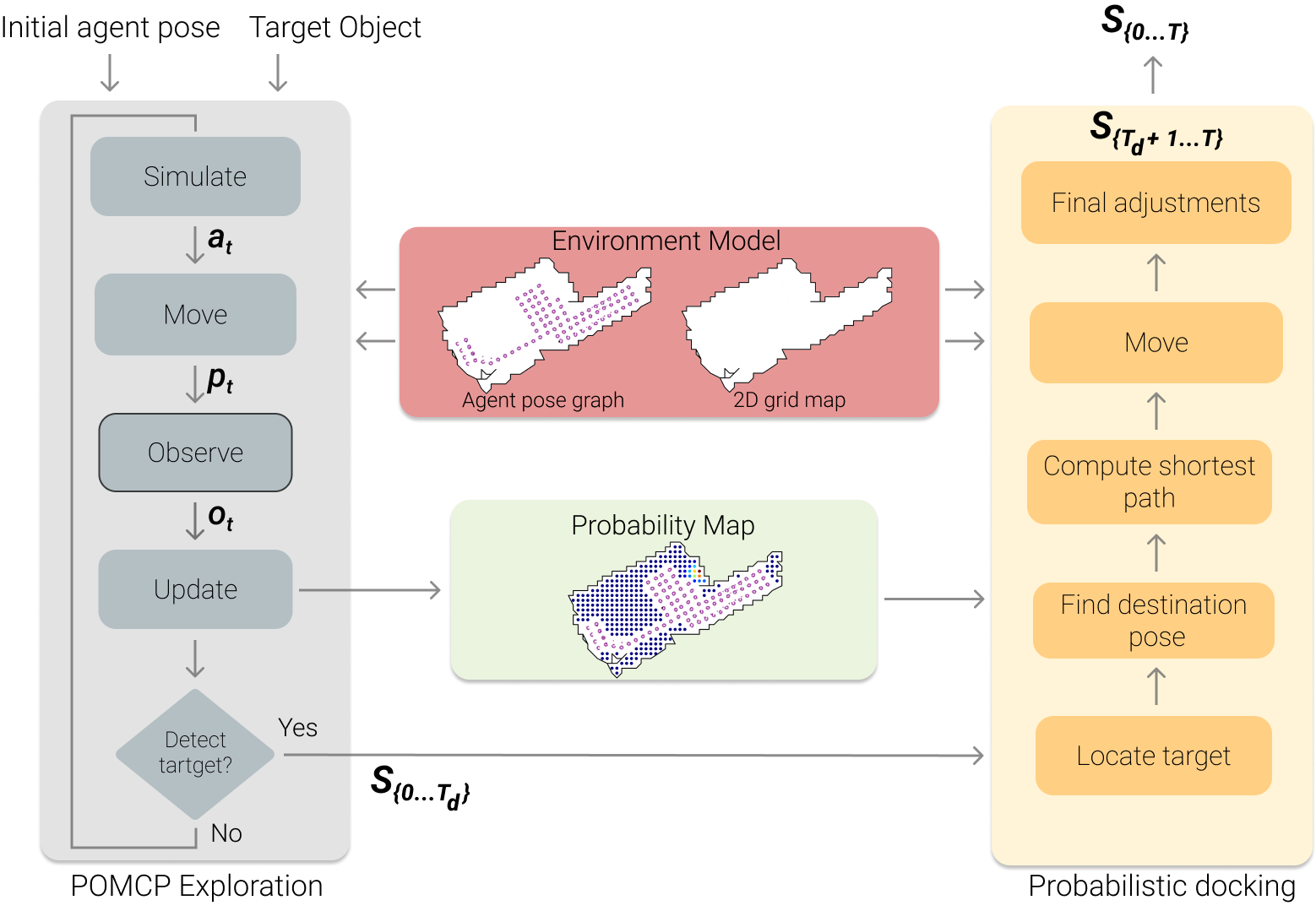}%
  \caption{Overall architecture of our proposed method \methname. The red box represents prior knowledge pushed into the POMCP module, the grey box represents the exploration strategy to detect the target object, the yellow box represents the probabilistic docking strategy to reach the destination pose and the green box represents the probability distribution over the locations. Math notation: state $s_t$, action $a_t$, pose $p_t$, observation $o_t$, POMCP state sequence $s_{\{0..T_d\}}$, docking state sequence $s_{\{T_{d+1}..T\}}$, complete state sequence $s_{\{ 0..T\}}$.}
   \label{fig:method}
\end{figure*}

A visual representation of our approach is shown in Fig.~\ref{fig:method}.
At each time step, we feed our model with the agent pose --\ie position and orientation-- in a known 2D map and a RGB-D frame given by a sensor acquisition. An off-the-shelf object detector is applied to the RGB image to identify the bounding box, if present, of the target object. The depth channel of the candidate target proposal is further exploited to obtain the candidate position in the floor map. We use this information to build a probability distribution over all the candidate locations of the target object. 
The policy is learnt online by Monte Carlo simulations, therefore it is general and easy to deploy in any environment.
The \emph{POMCP exploration} terminates when one location within the belief space exceeds a threshold.

Crucially, our approach exploits the model of the environment to consider the sensor's field of view and all the admissible moves of the agent in the area. For our active visual search scenario, such a model can be easily obtained by building a map of the environment to include the position of fixed elements, such as obstacles, walls or furniture.
Our motion policy explicitly exploits the knowledge of the environment for the visibility modelling, instead other RL-based strategies \cite{schmid2019iros,ye2019ral} implicitly encode such environment knowledge in a data-driven manner.

Once the exploration phase is over, the {\em probabilistic docking} module guides the agent to approach the target location --\ie the closest pose with a frontal-facing viewpoint to the target-- as quickly as possible. First, we estimate the shortest path~\cite{dijkstra1959note} on the graph of all possible robot poses, and then a path replanning is used to improve robustness.

With respect to our previous work~\cite{pomp2020bmvc}, the main contributions we make in this paper are:
\begin{itemize}
    \item a new strategy for improving the robustness with respect to false positives and miss-detection when using a real object detector in which we substitute the deterministic detection with a probabilistic one through a Bayesian inference considering a probability distribution over all possible object locations;
    \item a new strategy for the belief update of POMCP that allows us to lower the total path length of the exploration and increase the effectiveness with large state-space environments;
    \item a new approach for docking, considering the information gathered during the exploration to improve the robustness to the problems discussed above;
    \item a deeper experimental analysis and results discussion, providing results for all the scenarios of the Active Vision Dataset; and
    \item an extended description of the POMCP visual search method, with more mathematical details and discussion.  
\end{itemize}

\section{Related work}
\label{sec:soa}
The two main research topics related to this work are Active Visual Search and planning with Partially Observable MDPs. The main works of both topics are briefly surveyed in the following and original elements of our contribution with respect to the state-of-the-art highlighted.
\label{sec:soa:avs}

Active Visual Search, often referred to as Object Goal Navigation, is a specific task of Embodied AI research field. Embodied AI, which learns through interactions with the environment from an egocentric perspective, is an emerging field of study. Within this field, AVS is a task focused on detecting and approaching a specific object~\cite{batra2020objectnav}.

Early approaches exploit intermediate objects --\eg the relation between a sofa and a television-- to restrict the search area for the target object.
Although intermediate objects are usually easier to detect because of their size, their spatial relation \wrt the target may be not systematic. 
A probabilistic approach is proposed in~\cite{Kunze2014}, where the likelihood of the target increases when objects which are expected to be co-occurring are detected.

AVS with deep learning is viable using Deep Reinforcement Learning techniques~\cite{schmid2019iros,ye2019ral,han2019active}, where visual neural embeddings are often exploited for action policy training. Han \etal~\cite{han2019active} proposed a deep Q-network (DQN) where the agent state is given by CNN features describing the current RGB observation and the bounding box of the detected object. However, this work assumes that the object must be detected initially. To address the search task, EAT~\cite{schmid2019iros} performs feature extraction from the current RGB observation, and the candidate target crop generated by a region proposal network (RPN). The features are then fed into the Action Policy network. 
Similarly, GAPLE~\cite{ye2019ral} uses deep visual features enriched by 3D information, from the depth channel, for policy learning.
Although GAPLE claims to be generalised, expensive training is the cost to pay as GAPLE is trained with 100 scenes rendered using a simulator House3D based on the synthetic SUNCG dataset. 
This limitation is shared with other approaches that learn optimal policies using Asynchronous Advantage Actor Critic (A3C) algorithm~\cite{mirowskilearning}, Long Short Term Memories (LSTM) architectures~\cite{mousavian2019visual}, and Transformer networks coupled with deep Q-Learning~\cite{fang2019scene}.

Recent efforts from the community include also benchmarking the AVS task. Challenges including Habitat ObjectNav~\cite{habitatchallenge2022} encourage methods for enabling an agent initialised at a random starting pose in an \textit{unknown} environment to find a given instance of an object category using only sensory inputs to navigate, where large-scale datasets of 3D real-world spaces with densely annotated semantics are also made available to facilitate training and testing the models~\cite{yadav2022habitat3dsem}. As the scene map is unknown, most methods aim to learn the policy by joining the objectives of both semantic exploration and object search~\cite{chaplot2020object}. On top of such semantic exploration, the perception skills in terms of where to look and the navigation can be further disentangled~\cite{ramakrishnan2022poni} for an improved success rate. Moreover, spatial relations among objects have also been formulated as graphs and embedded via Graph Convolutional Networks to guide the navigation policy~\cite{Kiran2022srg}, where external commonsense knowledge has also shown advantages for the object localisation via spatial graph learning~\cite{giuliari2022spatial}.    

In general, RL-based strategies are dependent on training with a large amount of data in order to encode the environmental modelling and motion policy. Differently, our proposed POMCP-based method makes explicit use of the available scene knowledge and performs efficient planning for the agent's path online without additional offline training.

\subsection{Monte Carlo Planning}
\label{sec:soa:pomcp}
As for optimal policy computation, \emph{Partially Observable Markov Decision Processes} (POMDPs) are a popular framework for representing dynamical processes in uncertain environments and solving related sequential decision making problems~\cite{Kaelbling1998}.
Computing exact solutions for non-trivial POMDPs is often computationally intractable~\cite{Papadimitriou1987}, but in the recent years impressive progress was made to develop approximate solvers. One of the most recent and efficient strategies for solving POMDPs in an approximate way is \emph{Monte Carlo Tree Search} (MCTS) ~\cite{Thrun2000,Coulom2006,Browne2012}. The main advantage of using MCTS for solving POMDPs is scalability. 
MCTS-based strategies compute the policy online, \ie only for the specific states (or beliefs, in case of partially observable environments) the agent visits in its trajectories. This is fundamental in domains with very large state spaces and in partially observable environments where the dimension of the belief space is infinite, since beliefs are probability distributions over states. In MCTS system states are represented as nodes of a tree, and actions/observations as edges. Monte Carlo simulations are performed to generate the tree using specific action selection strategies, such as the algorithm called Upper Confidence bounds applied to Trees (UCT)~\cite{Kocsis2006}, that efficiently balances exploration and exploitation. 

The most influential solver for POMDPs which takes advantage of MCTS is \emph{Partially Observable Monte Carlo Planning (POMCP)}~\cite{Silver2010} which combines a particle filter representation of the belief, a MCTS-based strategy for computing action Q-values, and an efficient way to update the agent's belief. Several extensions of POMCP have been realised. BA-POMCP~\cite{Katt2017} extends POMCP to Bayesian Adaptive POMDPs, allowing the model of the environment to be learned during execution. A version of POMCP for scalable planning in multiagent POMDPs is presented in~\cite{Amato2015}, it introduced model learning in POMDPs considering also the uncertainty about model parameters in the belief. A scalable extension of POMCP for dealing with cost constraints is presented in~\cite{Lee2018}. Very recent work focused on the introduction of prior knowledge about the environment and the policy in POMCP. In~\cite{Castellini2019a} known state-variable relationships are used to improve the performance of POMCP. In~\cite{Mazzi2021AAMAS} logical rules representing parts of the POMCP policy are generated using Satisfiability Modulo Theory (SMT) to improve the explainability of the policy and identify anomalous action selections due to wrong parameter tuning. Again, in the research line of merging probabilistic planning and symbolic approaches, \cite{Mazzi2021ICAPS} allows to generate shields based on logical rules to improve the safety of POMCP. The technique has been further improved in~\cite{Mazzi22AAMAS,Meli23AAMAS} where active approaches and methods based on Inductive Learning of Answer Set Programs are used to learn the logic rules.

Applications of POMCP can be found in several domains. A few of them are related to the exploration of partially known environments \cite{Lauri2016} and the find-and-follow of people \cite{Goldhoorn2014} with robots. Others \cite{castelliniEAAI2021,zuccotto20222} concern robot navigation using only POMCP or hierarchical methods approaches with POMCP for high-level control and neural networks for low-level control.  Popular MCTS-based approaches have been recently used also for developing agents with superhuman performance in the game of Go \cite{Silver2016,Silver2018}. The approach proposed in this work differentiates from all 
works mentioned above because it specialises POMCP to the AVS domain and introduces methodological improvements to belief update, probabilistic detection of objects and docking that are not present in the literature. To the best of our knowledge, the only works available in the literature about AVS with POMCP are \cite{pomp2020bmvc,Giuliari2021}. The differences with these works are substantial since \cite{pomp2020bmvc} uses standard belief update, assumes exact object detection and employs a na\"ive docking procedure, while \cite{Giuliari2021} focuses on completely unknown environments.

\section{Method}
\label{sec:method}
We consider an agent navigating through known environment with the goal of locating and approaching a specific object. The agent explores the environment to identify the target object, determines its location on the map and then moves closer to it.

To be coherent with the related literature, the agent’s \emph{pose}\footnote{Here with pose we mean the 2D robot pose, \ie position and orientation. We do not consider here complex kinematics related to the agent structure (\eg if a humanoid robot is used).} at time step $t$ is $p_t=\{x_t,y_t,\theta_t\}$, where $x$ and $y$ are the coordinates on the floor plane, and $\theta$ is the orientation.
At each time step the agent takes an action $a_t$ from a predefined set $A$: specifically, the agent can \verb|move_forward|, \verb|move_backward|, \verb|rotate_clockwise|, \verb|rotate_counter_clockwise|. Rotations are defined by a fixed angle.

When the agent reaches a new pose $p_t$, it receives an observation which is the output of an object detector applied to the image acquired by an RGB-D camera.
We model the search space as a grid map (see Figure~\ref{fig:intro_figure}(b)), in which each cell can be either:
\textit{(i)} ``\textit{visual occlusion}'', if the cell is occupied by obstacles, such as a wall or a piece of furniture, that prevent the agent to see through;
\textit{(ii)} ``\textit{empty}'', if the agent is allowed to enter the cell and thus no objects can be located in there; or 
\textit{(iii)} ``\textit{candidate}'', if none of the above, thus the cell is a possible object location for the target object.

\subsection{Partially Observable Markov Decision Processes}
We formulate the AVS problem as a Partially Observable Markov Decision Process (POMDP), which is a standard framework for modeling sequential decision processes under uncertainty in dynamical environments \cite{Kaelbling1998}. 
A POMDP is a tuple $(S,A,O,T,Z,R,\gamma)$, where $S$ is a finite set of partially observable states, $A$ is a finite set of actions, $Z$ is a finite set of observations, $T$:~$S\times A \rightarrow \Pi(S)$ is the \textit{state-transition model}, $O$:~$S\times A \rightarrow \Pi(Z)$ is the \textit{observation model},  $R$:~$S \times A \rightarrow \R$ is the reward function and $\gamma \in [0,1)$ is a discount factor.
Agents operating POMDPs aim to maximise their expected total discounted reward 
$E[\sum_{t=0}^{\infty} \gamma^t R(s_t,a_t)]$, by choosing the best action $a_t$ in each state $s_t$, where $t$ is the time instant; $\gamma$ reduces the weight of distant rewards and ensures the (infinite) sum's convergence. The partial observability of the state is modelled by considering at each time-step a probability distribution over all the states, called the \textit{belief} $B$. 
POMDP solvers are algorithms that compute, in an exact or approximate way, a \textit{policy} for POMDPs, namely a function $\pi$:~$B \rightarrow A$ that maps beliefs to actions. 

\subsection{Partially Observable Monte Carlo Planning} 
\label{sec:pomdp}
Partially Observable Monte Carlo Planning (POMCP)~\cite{Silver2010} is an online Monte-Carlo based solver for POMDPs. It uses Monte-Carlo Tree Search (MCTS) for selecting, at each time step, an action which approximates the optimal one.
Given the current belief, represented by an unweighted particle filter, the Monte Carlo tree is generated by performing a certain number of simulation from the current belief. 
These simulations generate, in an efficient way, estimates of the Q-values of all actions from the current belief. The action with the highest estimated Q-value is selected and performed in the real environment. A big advantage of POMCP is that it enables to scale to large state spaces because it never represents the complete policy but it generates only the part of the policy related to the belief states actually seen during the plan execution. Moreover, the local policy approximation is generated online using a simulator of the environment, namely a function that given the current state and an action provides the new state and an observation according to the POMDP transition and observation models.
An important step of the POMCP algorithm is belief update. Every time the agent moves in the real world, the belief is updated by POMCP to reflect the new information acquired by observing the environment. 
 After the agent has performed all simulations and taken a single action $a$ in the real world it receives an observation $o$. The new belief is initialised to the particle filter of the node of the tree reached selecting the branch with action $a$ and then observation $o$ in the tree from the current root node.

\subsection{Exploration, localisation and approach}
The methodology here proposed is a specialisation of POMCP for the AVS problem. It is based on three main elements, defined in the following, that are used altogether by POMCP to perform the search of an object in the environment. We assume that $n$ is the number of possible poses 
that the agent can assume in the environment,
$m$ is the number of objects in the environment, and $k$ is the number of locations in which each object can be positioned.\\
\textit{(i)} The first element of the proposed framework is a \emph{pose graph} $\mathcal{G}$ in which nodes represent the $n$ possible poses of the agent and edges connect only poses reachable by the agent with a single action. Thus, $\mathcal{G}$ is used to constrain the actions that cannot be performed in the real and simulated world.
\textit{(ii)} The second element is the set $\mathcal{H} = \{ 1, \ldots, k \}$ of all possible indices of locations that each object can take in the environment.
Each index in $\mathcal{H}$ corresponds to a specific position in the topology of the environment where the search is made.
\textit{(iii)} The third element is a matrix of object observability $\mathbf{L} = (l_{i,j}) \in \{0,1\}^{n \times k}$, where $l_{i,j}=1$ if the location $j$ is visible from pose $i$ --\ie location $j$ is in the field of view of the agent positioned in $i$. Matrix $\mathbf{L}$ can be deterministically derived from $\mathcal{G}$ and $\mathcal{H}$ by a visibility function $f_L$ which computes the visibility of each object from each agent pose, considering the physical properties of the environment. This matrix is used in the observation model employed in the simulations. Namely, the observation model returns 1 if the target object is observed from the current pose, 0 otherwise. More formally, it returns 1 if the agent is in position $\hat{i} \in \mathcal{G}$ and the target object is in a position $\hat{j}\in \mathcal{H}$ for which $l_{\hat{i},\hat{j}}=1$. Notice that the position of the target object in each specific simulation is known because it is defined in the particle sampled at the beginning of the simulation. On the other hand, observations in the real world are based on the object detector. 
Both for the real and simulated world, we give a positive reward if the object is observed; 
otherwise, a negative reward is provided (corresponding to the energy spent to perform the movement) and the POMCP-based search is continued.

To prevent the agent to visit the same poses more than once, the agent maintains an internal memory vector that collects all the poses already visited during the current run. Every time the agent re-visits a pose it receives a high negative reward.
After every step in the real world, the agent receives from the object detector an observed value $1$ if the target object has been observed, $0$ otherwise.

The belief of the agent at each time step is an approximated probability distribution over all the candidate object locations in the environment, that represents the POMCP hidden state.
If the object is not observed within a fixed amount of moves, the method terminates and reports a search failure.

\subsubsection{Belief update} 
\label{sec:method:belief}

In our original formulation of POMP \cite{pomp2020bmvc}, belief is updated using the standard POMCP strategy. 

A problem with this approach is related to the cardinality of our state space. In AVS, the state space describes both the agent's pose and the target's location. Because the object can be in any location, and the number of simulations is limited, it may happen that some states are not considered during the simulation phase and can only be recovered during reinvigoration. If they are not recovered, they are removed from the belief and cannot be recovered anymore, even if they are valid positions.
Another issue with this approach is that, during reinvigoration, the new particles are sampled from the previous belief. This creates a feedback loop in which particles that survive the belief update, have a higher chance of being chosen during reinvigoration. 
Thus, in situations where the number of simulations is limited, it is possible for the belief to become confined to a sub-space within the state space.

In \methname, we change the belief update and resampling procedure to overcome these issues. The belief is initially generated by sampling particles from a uniform distribution over all states --\ie candidate object locations. An auxiliary variable $pp$ stores the current list of object locations not been observed yet, where  $pp=\{ j \in \mathcal{H} \ | \ j \ \text{not yet observed}\}$.
The set is initialised as $pp=\mathcal{H}$. At each time step, the agent acquires observations about the locations within the current FOV through the object detector, and it updates $pp$ removing the observed positions that do not contain the searched object. The new belief is sampled from a uniform distribution over locations satisfying the $pp$ constraint --namely, having the searched object in positions belonging to $\mathcal{H} \setminus pp$. This way to update the belief is beneficial in terms of performance, as shown in our experiments below.

\subsubsection{Probabilistic Detection}

We equipped our agent with the \emph{Target Driven Instance Detector} (TDID) presented in~\cite{ammirato2018target}, an architecture designed to recognise and classify specific instances of object classes. 
Given an image, TDID returns a list of coordinates representing the associated bounding box (if any), a score $s \in [0, 1]$ and the corresponding class $c$. In our work, we consider only detections with an associated score greater than $0.9$. Moreover, given the rate of \textit{TP} (True Positive), \textit{FP} (False Positive) and \textit{FN} (False Negative), we define:
\begin{equation*}
  Precision = \frac{TP}{ TP + FP} \hspace{.1\linewidth}
  Recall = \frac{TP}{ TP + FN} 
\end{equation*}

\noindent On top of that, we define the $F_1$ score as: 
\begin{equation*}
  F_{1}\text{-score} = 2 \; \frac{Precision \times Recall}{Precision + Recall} \; ,
\end{equation*}
where $F_1 \in \left[0,1\right]$ can be interpreted as the harmonic mean of $Precision$ and $Recall$.


\begin{figure}[t]
\begin{center}
        \resizebox{\linewidth}{!}{
	\begin{tabular}{@{}c@{}c@{}}
		\includegraphics[width=0.33\textwidth]{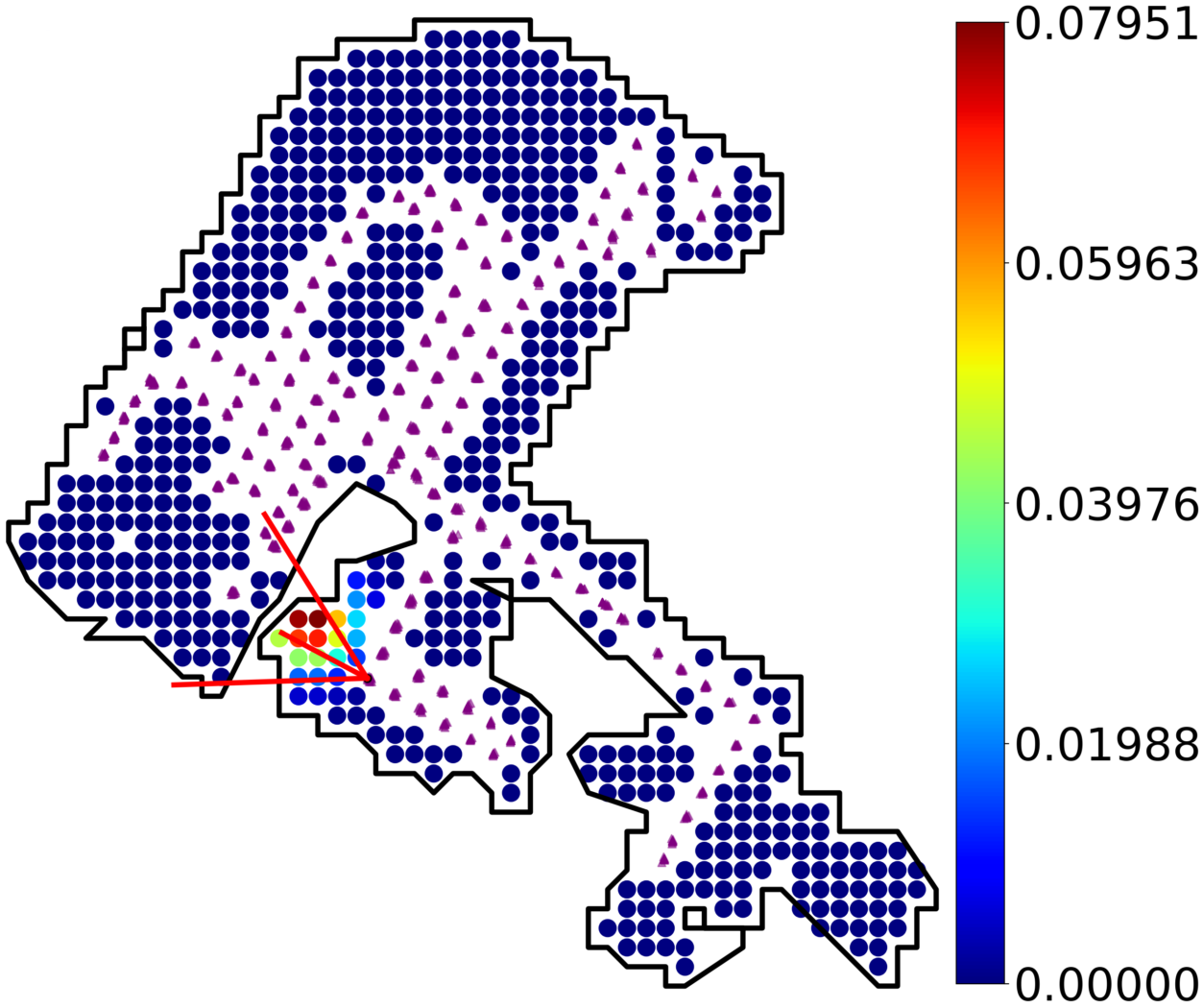}&
		\includegraphics[width=0.33\textwidth]{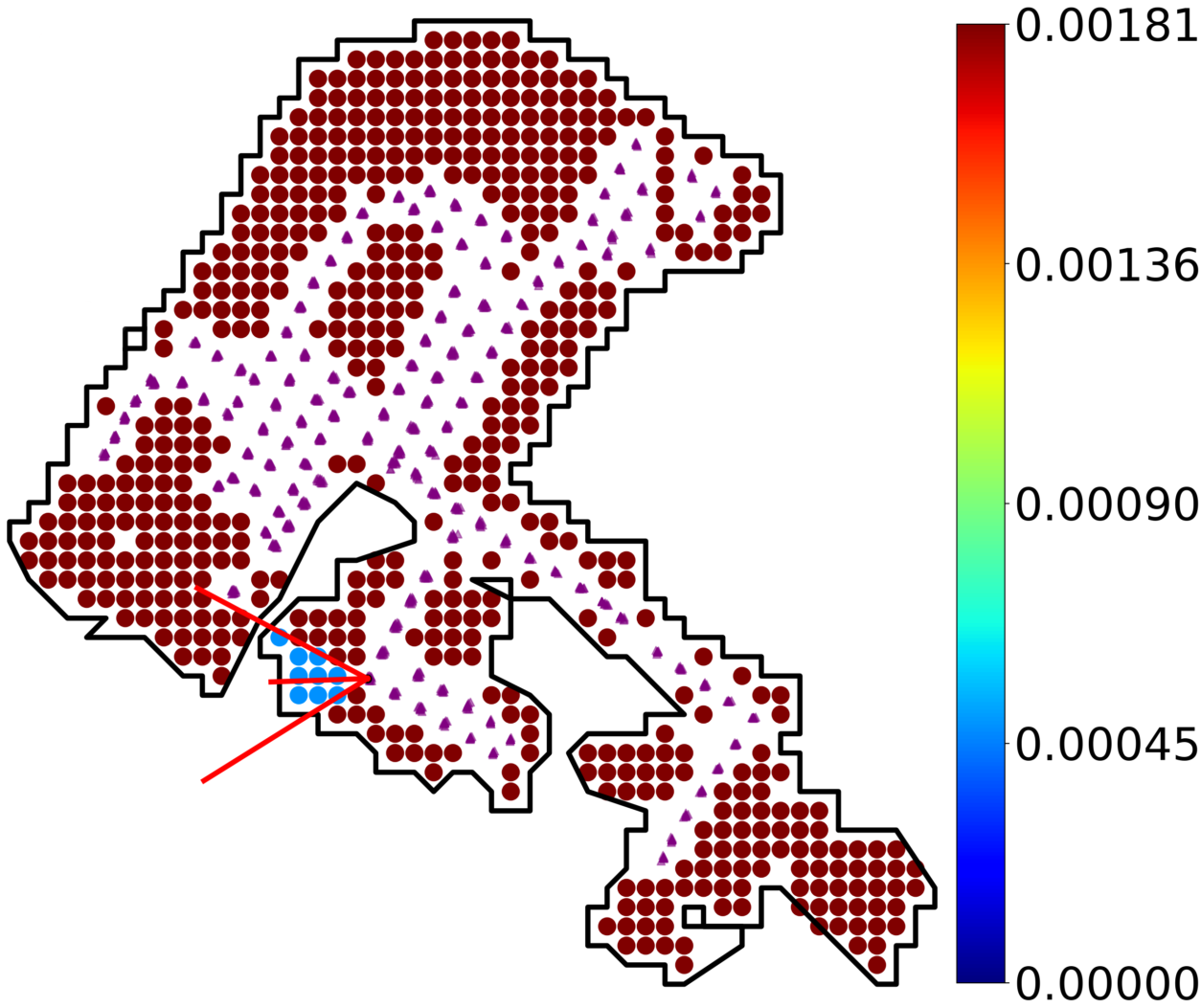}\\
		(a) Object inside FOV & (b) Object outside FOV
   	\end{tabular}}
\caption{The two cases considered when creating the vector $D$. Example derived from Home\_003\_2. In case (a) the objective is to determine the location of the object and assign probabilities in the form of a multivariate normal distribution. In (b), we assign low probabilities to the locations inside the FOV, and high probabilities to the locations outside it. Note: we assign different scales to the colorbar for ease of visualisation.}
\label{fig:inside_outside_cases}
\extralabel{fig:inside_outside_cases:a}{(a)}
\extralabel{fig:inside_outside_cases:b}{(b)}
\end{center}
\end{figure}

\begin{figure*}[t]
\begin{center}
	\begin{tabular}{@{}c@{}c@{}c@{}}
		\includegraphics[width=0.33\textwidth]{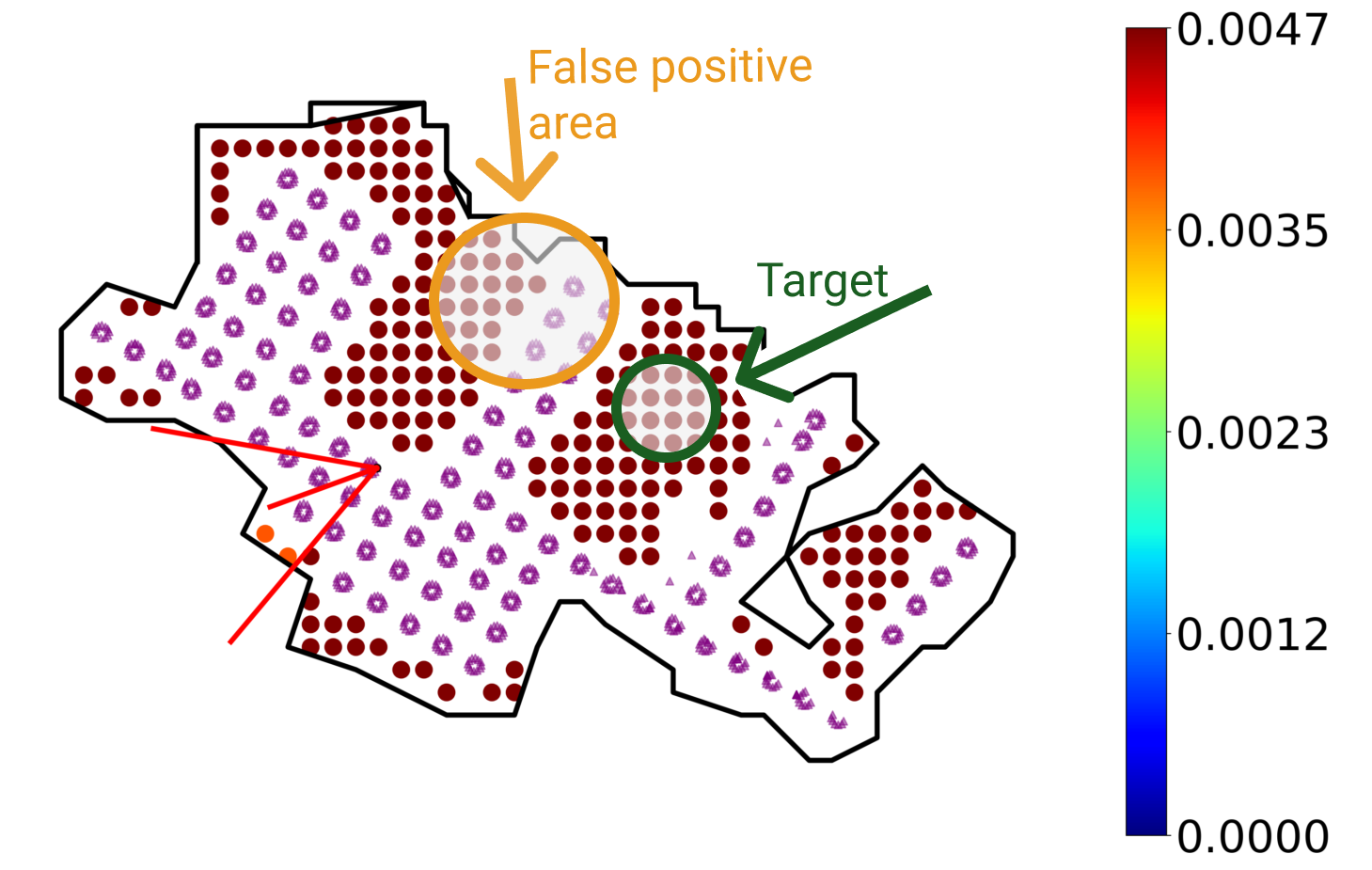}&
            \includegraphics[width=0.33\textwidth]{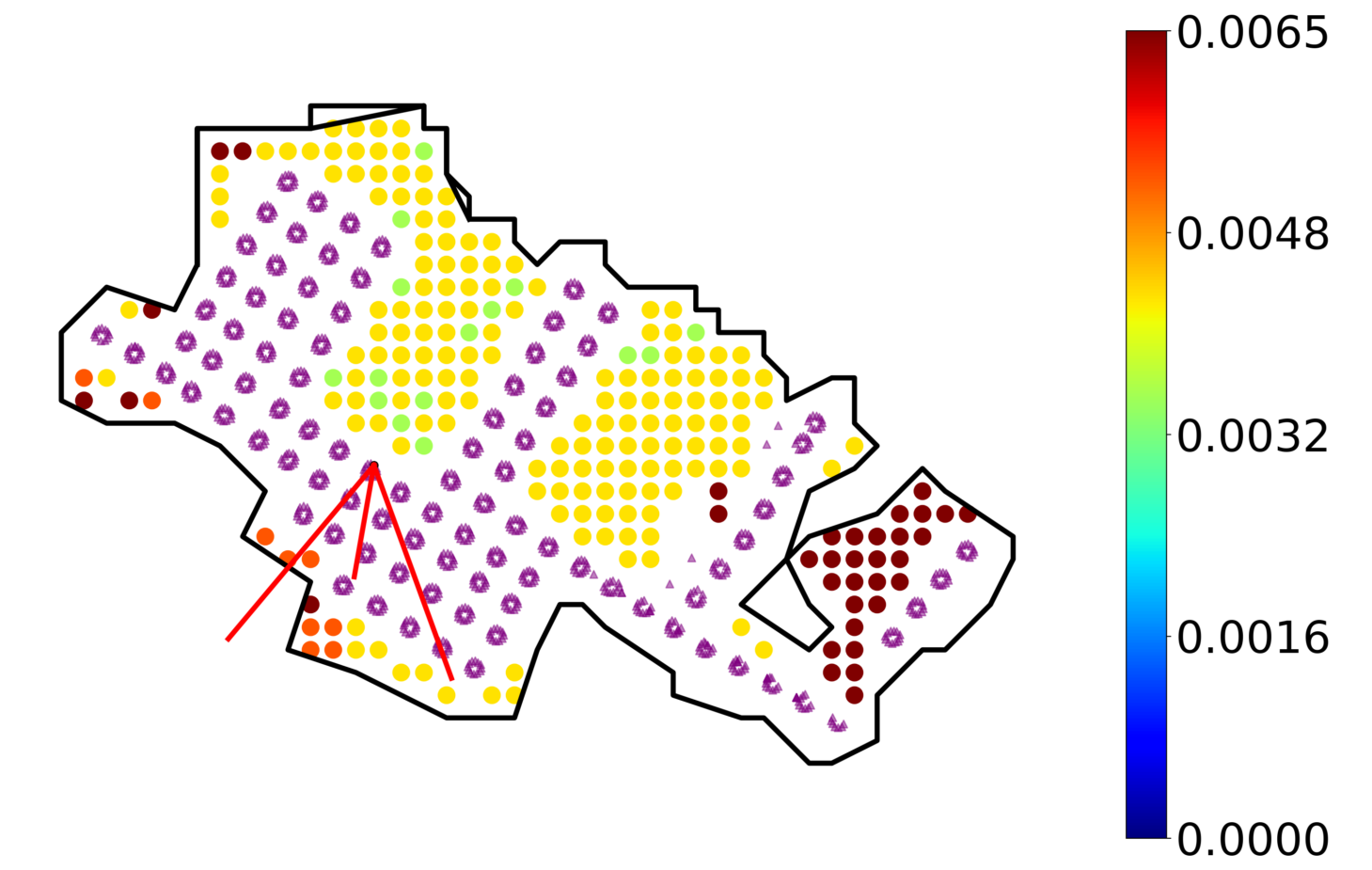}&
		\includegraphics[width=0.33\textwidth]{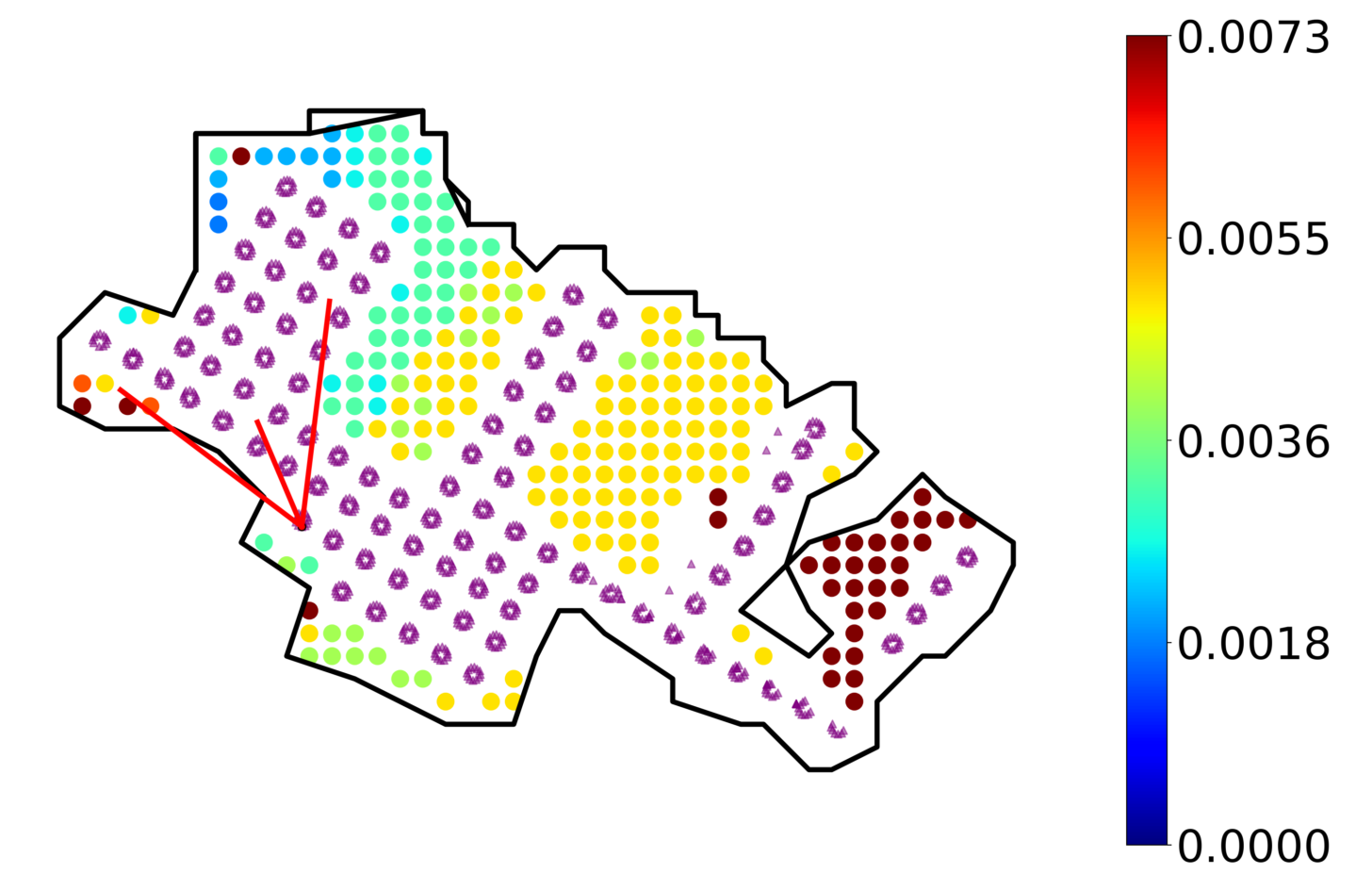}\\
		(a) Step 1  & (b) Step 10 & (c) Step 20
   	\end{tabular}
        \begin{tabular}{@{}c@{}c@{}c@{}}
		\includegraphics[width=0.33\textwidth]{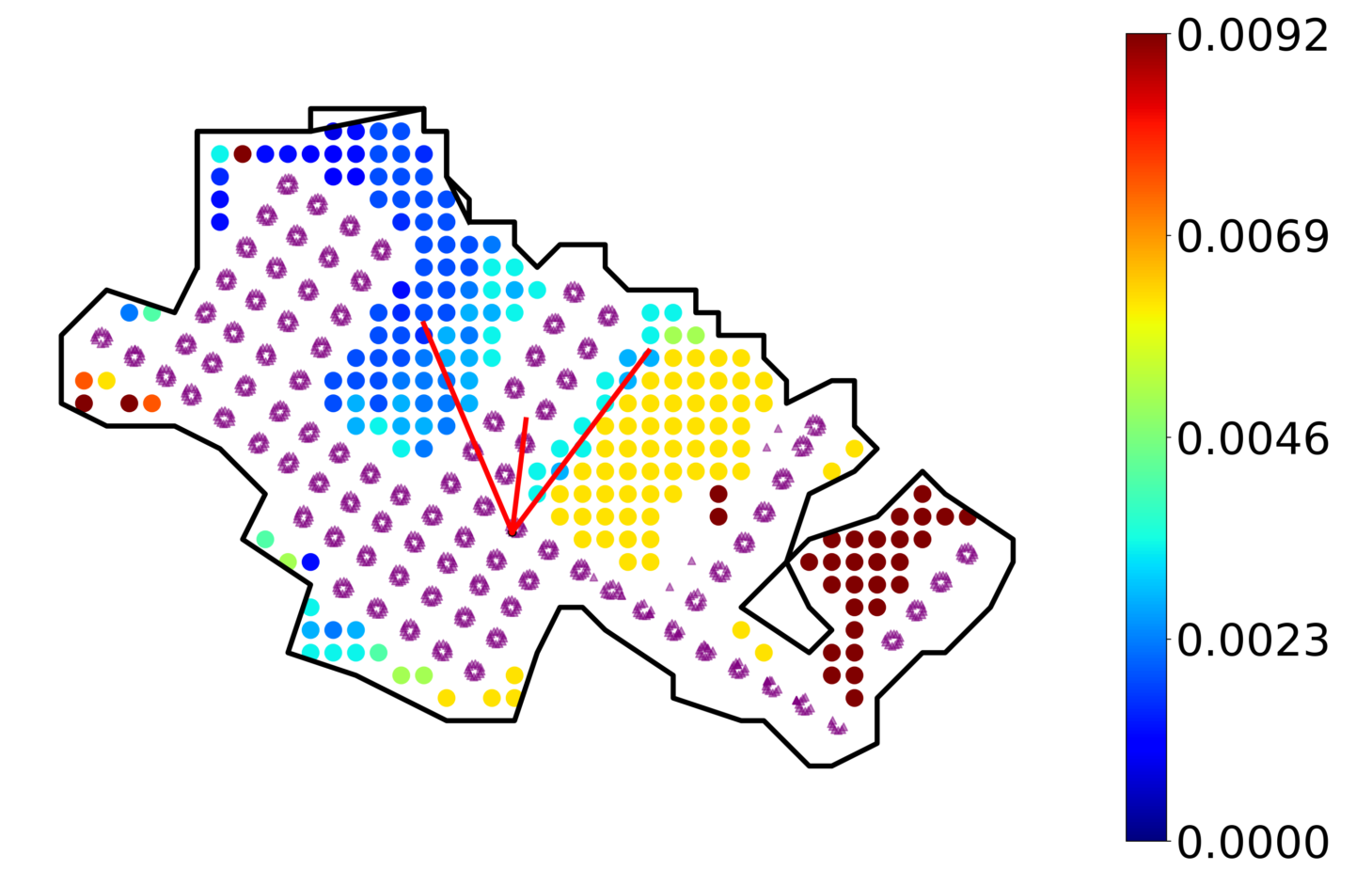}&
		\includegraphics[width=0.33\textwidth]{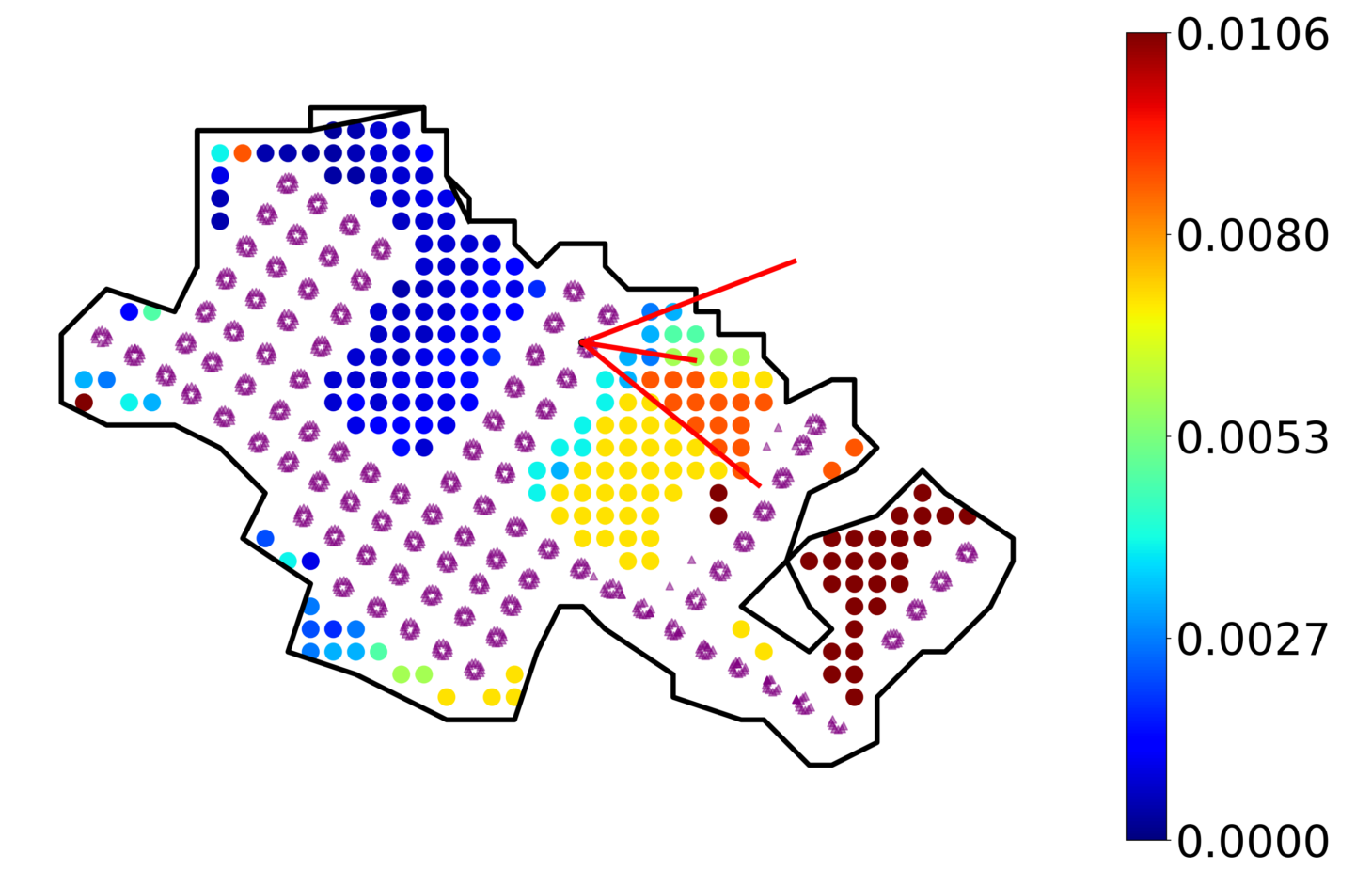}& \includegraphics[width=0.33\textwidth]{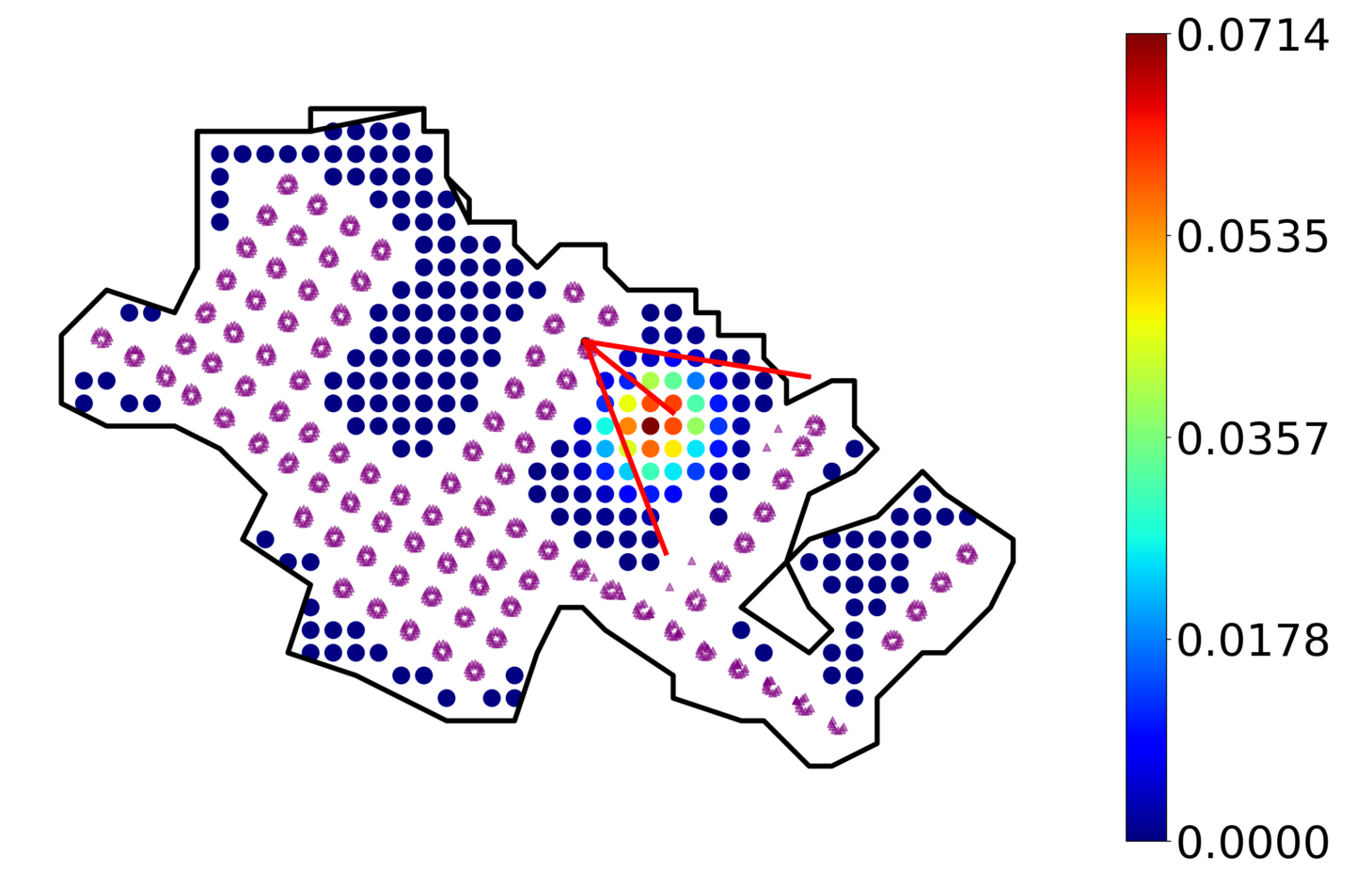}\\
		(d) Step 33  & (e) Step 50 & (f) Step 51 
   	\end{tabular}
\end{center}
\caption{Experiment inside Home\_016\_1 using the proposed approach \methname. In step (a) we initialise the agent in the environment; we highlight the target position and a false positive area. From step (b) to (c) the robot explores the top area; in step (d) we show the robustness of our approach to a false positive; finally, in step (e) we identify high probable locations, locating the target in step (f).}
\label{fig:step_by_step}
\extralabel{fig:step_by_step:a}{(a)}
\extralabel{fig:step_by_step:b}{(b)}
\extralabel{fig:step_by_step:c}{(c)}
\extralabel{fig:step_by_step:d}{(d)}
\extralabel{fig:step_by_step:e}{(e)}
\extralabel{fig:step_by_step:f}{(f)}
\end{figure*}

In POMP~\cite{pomp2020bmvc} the planner terminates the exploration phase  when the object detector identifies the target object inside the FOV. In this way, false positives of the object detector produce wrong terminations of the exploration in positions where the object is not present.
In \methname we aim to reduce the impact of this problem.
We first define a vector $\mathcal{D}=\{d_1,\ldots, d_k\}$ for the probabilities to have the target object in location $j$ considering only the observation at the current time step. In other words, at each step in the real environment, we reset all $d_j$ with $j=1,\ldots,k$ considering only the current observation. If the object is found by the object detector inside the current FOV, then we set $p_j=0$ in locations $j$ outside the FOV and we set the probabilities $p_j$ of locations inside the FOV according to a  multivariate normal distribution with mean in the location $j$ where the object is localised by the detector (see Fig.~\ref{fig:inside_outside_cases:a}). 
If the object is not found by the object detector inside the FOV, then we set $d_j=F_1$ in locations $j$ inside the FOV and $d_j=1 - F_1$ in locations $j$ outside the FOV (see Fig.~\ref{fig:inside_outside_cases:b}). 
In both cases, we normalise $\mathcal{D}$ so that $\sum_{j=1}^k d_j=1$. 
Notice that $F_1$ is class specific --\ie it accounts for the performance of the  object detector for the specific object class.
As a second auxiliary data structure, we define a vector $\mathcal{R} = \{ p_1, \ldots, p_k \}$ of probabilities to have the target object in location $j$ considering the whole history of observations --\ie this represents a global probability using information also from previous steps. In particular, value $p_j$ is updated at each time step $t$ according to the following rule:
\begin{equation}
     p_j^t = \frac{p_j^{t-1} \cdot d_j^t }{\sum_{i=1}^k p_i^{t-1} \cdot d_i^t}
     \label{eqn:update_rule}
\end{equation}
for all $j \in \mathcal{H}$.
Finally, we define a threshold $\tau = \frac{c}{n}$, where $n$ is the number of candidate object locations, and $c \in \N$ is a constant that allows us to increase the confidence of our probabilistic detection. We terminate the POMCP exploration phase when in the FOV of the current pose we have an object location $j$ whose probability $p_j$ exceeds the threshold. More formally, the following exit condition must be verified: 
\begin{equation}
  (p_j \geq \tau) \land (L_{i,j} = 1).  
     \label{eqn:exit_condition}
\end{equation}

According to this procedure, if the object is not in the current FOV, we assume that it must be in some other location, thus we increase the corresponding probabilities. Instead, if it is in the current FOV, we increase the probabilities of the locations near the 3D position of the object and lower the other ones. 
Moreover, we do not rely merely on the object detector output, we rather accumulate knowledge over time leveraging the old and current state of the environment. In Fig.~\ref{fig:step_by_step} we report an episode in which we can appreciate the evolution of the probabilities inside an environment.

\vspace{.5em}
\subsubsection{Probabilistic Docking}

Given the object location $j\in \mathcal{H}$ satisfying the exit condition of Eq.~\ref{eqn:exit_condition}, we first identify the \emph{destination pose} --\ie the agent's pose $\hat{i}\in \mathcal{G}$ that is closest to the target location and points towards it.
Then we use the Dijkstra algorithm~\cite{dijkstra1959note} to compute the shortest path between the current pose $i\in \mathcal{G}$ and the estimated destination pose $\hat{i}\in \mathcal{G}$.

While the agent navigates towards the destination pose, the object detector is not used since we are confident enough that the target object is in location $j$.

This strategy achieves better performance than the \textit{Robust Visual Docking} introduced in~\cite{pomp2020bmvc}. A key distinction is that in Robust Visual Docking the object detector is used along the path, thus, in case of poor performing detectors, miss-detections and false positives can easily distract the agent from the final goal, having fatal consequences in the approaching phase.

\section{Experiments}
\label{sec:exp}

\begin{figure*}[!t]
\begin{center}
	\begin{tabular}{@{}c@{}c@{}c@{}c}
		\includegraphics[width=0.25\textwidth]{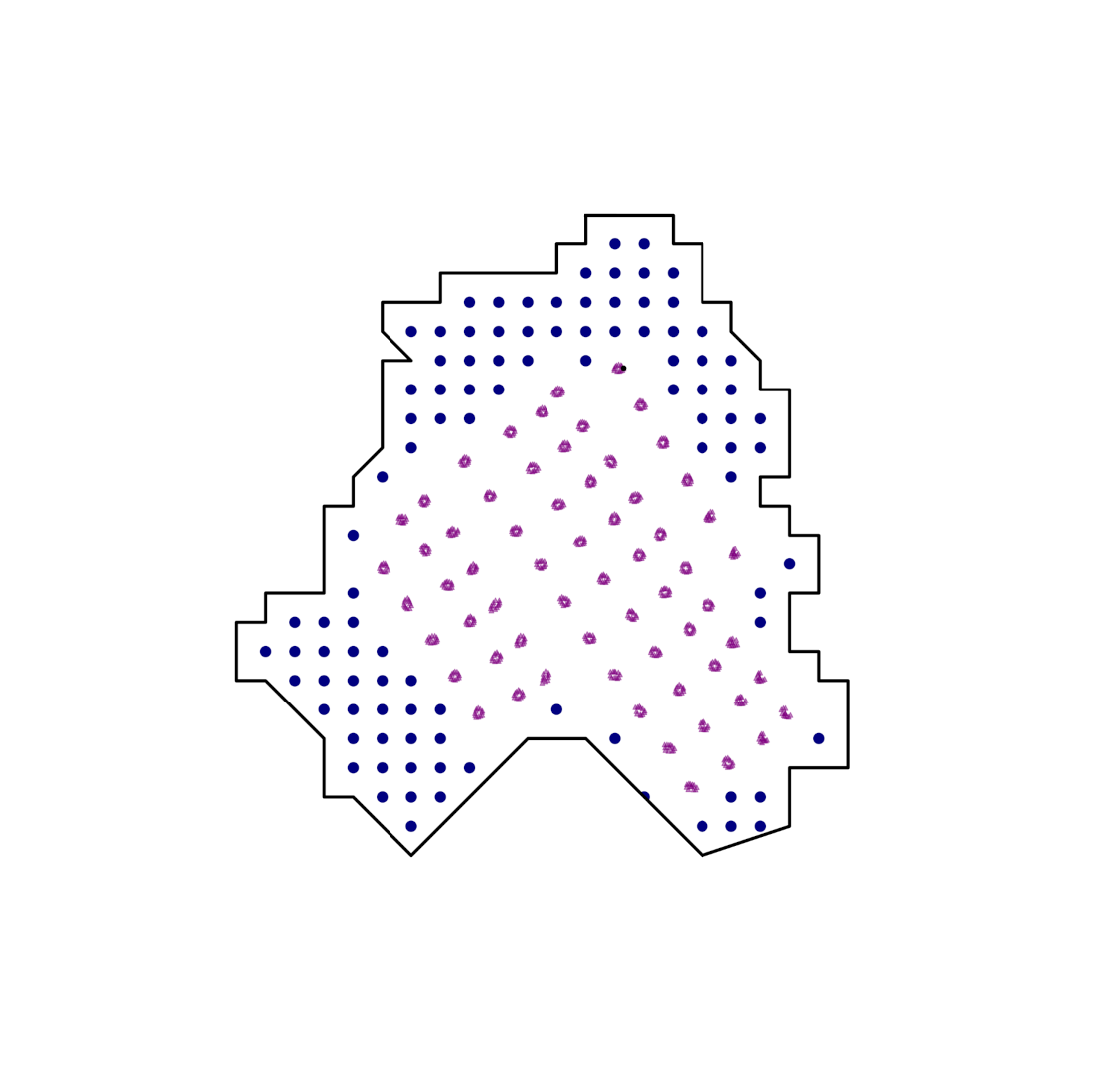}&
		\includegraphics[width=0.25\textwidth]{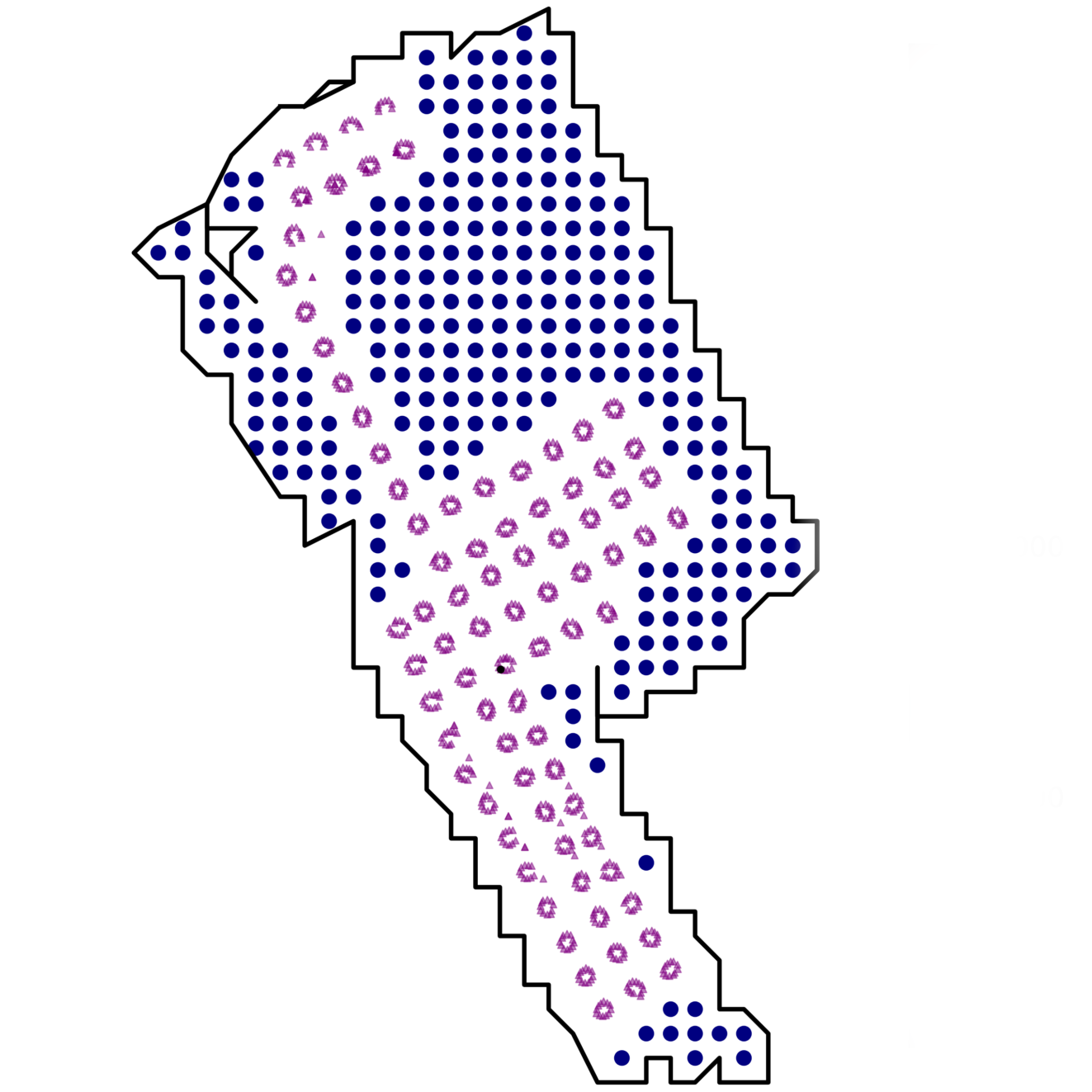}&
		\includegraphics[width=0.25\textwidth]{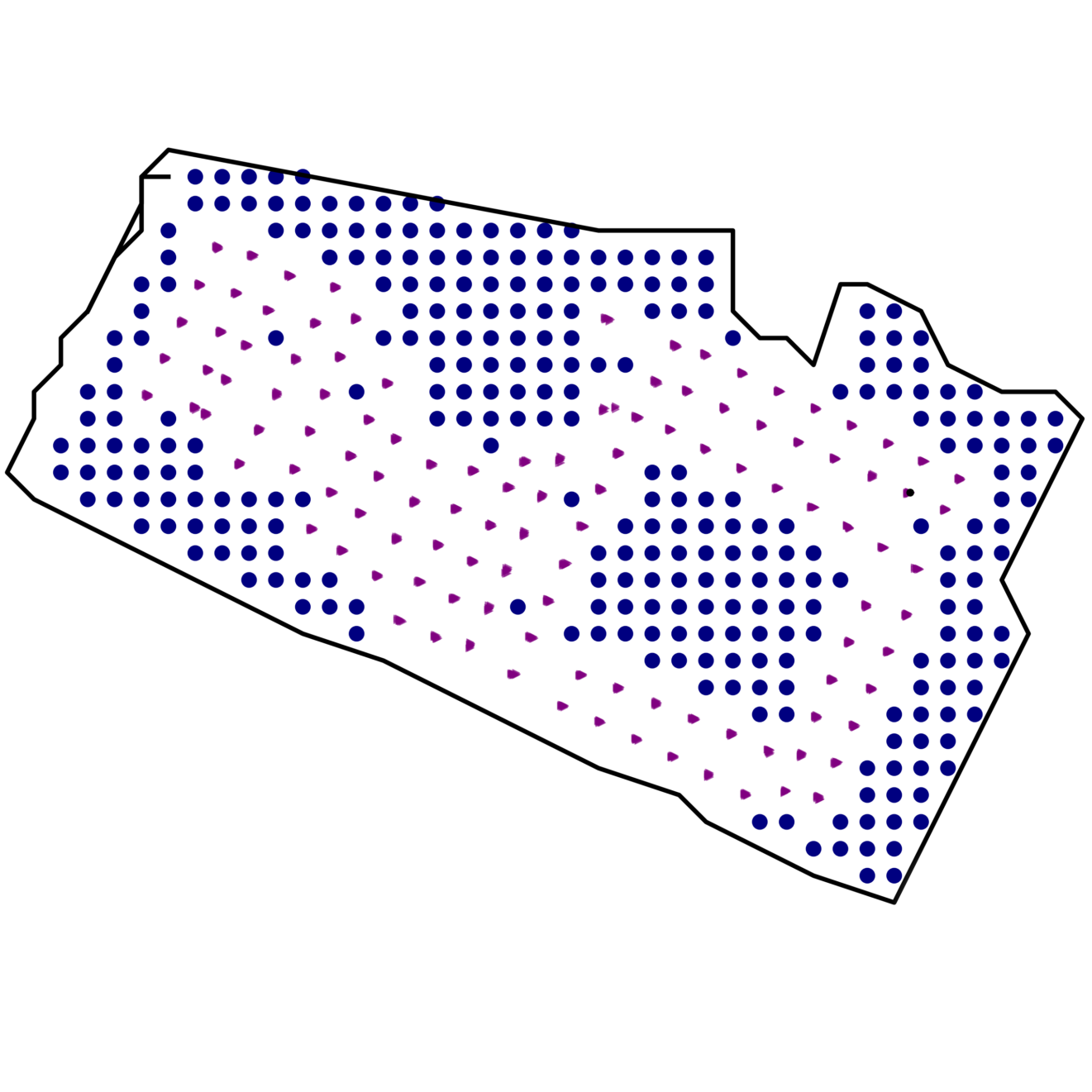}&
		\includegraphics[width=0.25\textwidth]{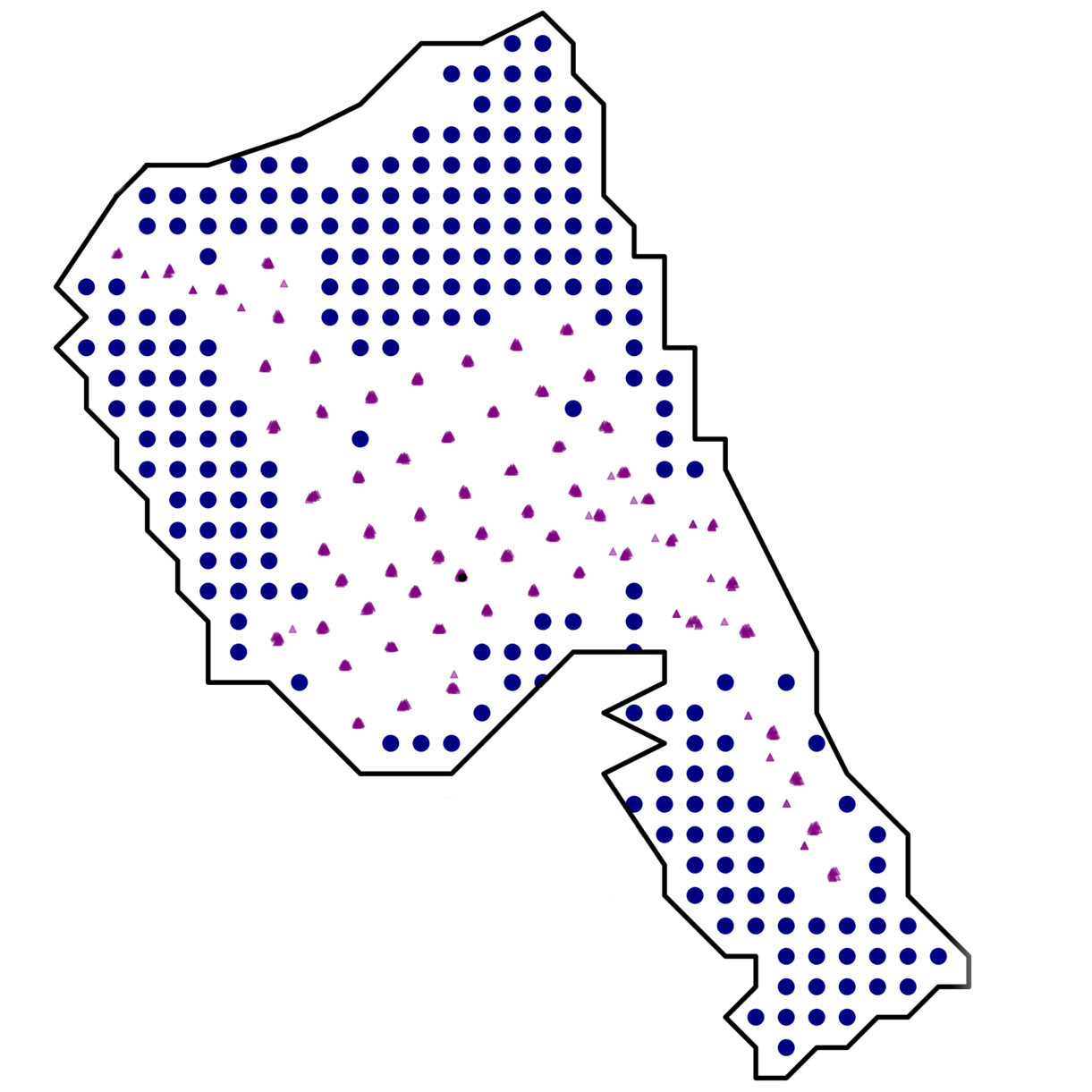}\\
		(a) Easy: Home\_005\_2 [97] & (b) Easy: Home\_015\_1 [265] & (c) Med: Home\_001\_2 [289] & (d) Med: Home\_014\_2 [243]\\
		\includegraphics[width=0.25\textwidth]{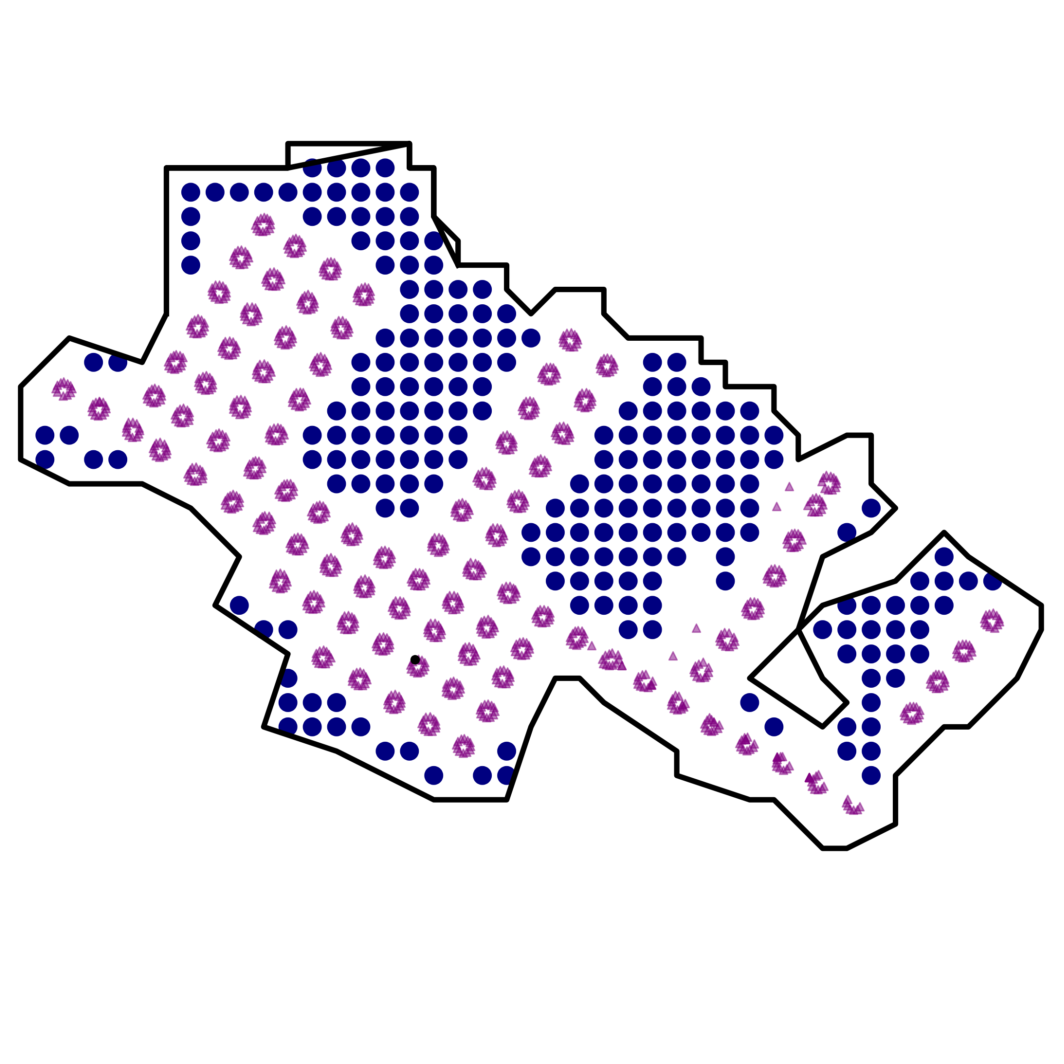}&
		\includegraphics[width=0.25\textwidth]{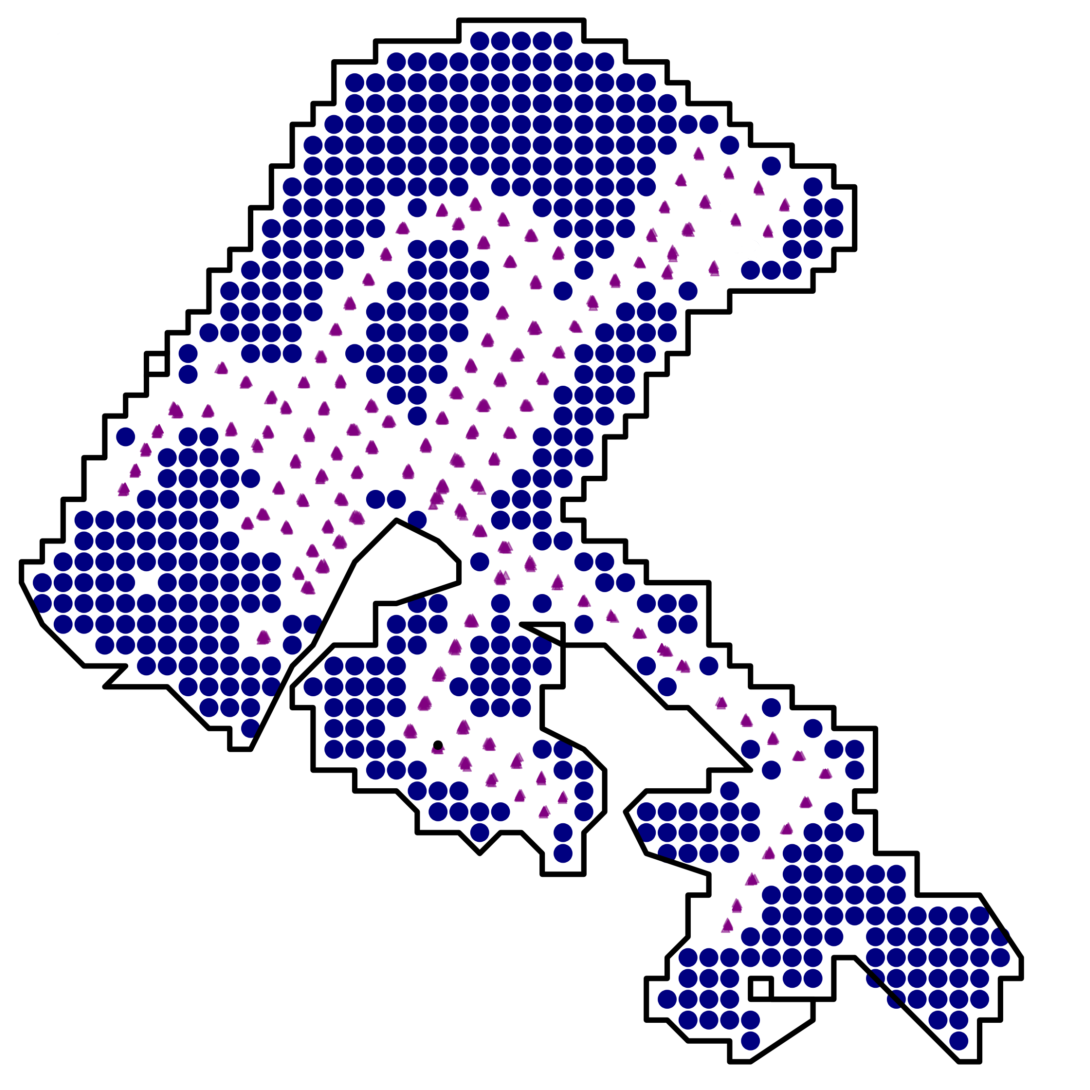}&
		\includegraphics[width=0.25\textwidth]{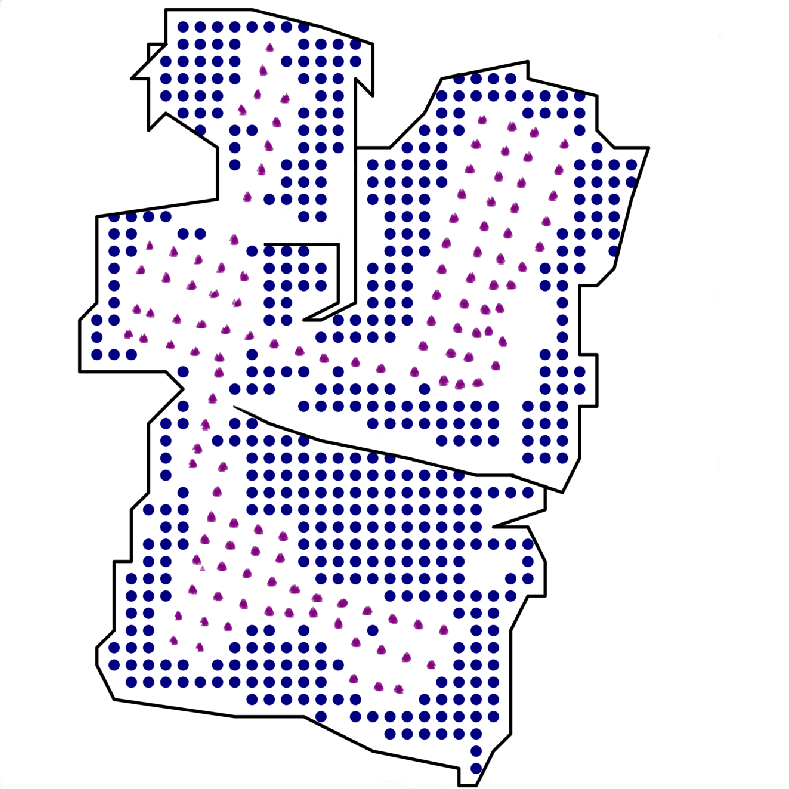}&
		\includegraphics[width=0.25\textwidth]{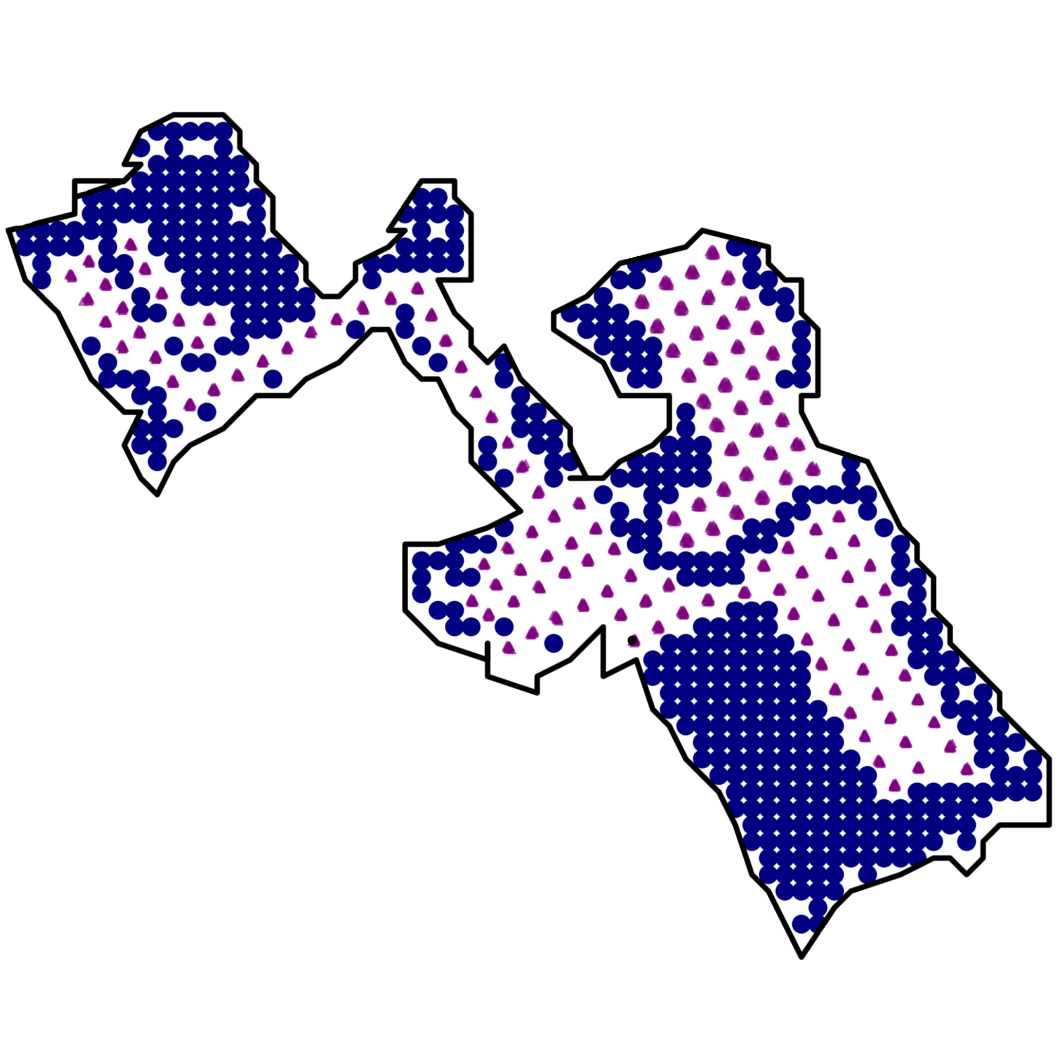}\\
		(e) Med: Home\_016\_1 [215] & (f) Hard: Home\_003\_2 [560] & (g) Hard: Home\_004\_2 [490] & (h) Hard: Home\_013\_1 [490]\\
   	\end{tabular}
\end{center}
\vspace{-0.3cm}
\caption{Corresponding 2D floor maps (not in scale) for the test scenes from AVBD of 3 different difficulty levels (as in \cite{schmid2019iros}).
For each environment, we report the name and, in parenthesis, the number of possible object locations.
As the difficulty increases, we can note an increment of possible object location and more difficult spatial layouts.}
\label{fig:home_maps}
\extralabel{fig:home_maps:a}{(a)}
\extralabel{fig:home_maps:b}{(b)}
\extralabel{fig:home_maps:c}{(c)}
\extralabel{fig:home_maps:d}{(d)}
\extralabel{fig:home_maps:e}{(e)}
\extralabel{fig:home_maps:f}{(f)}
\extralabel{fig:home_maps:g}{(g)}
\extralabel{fig:home_maps:h}{(h)}
\end{figure*}

\begin{figure*}[!t]
\begin{center}
	\begin{tabular}{@{}c@{}c@{}c@{}c}
		\includegraphics[width=0.23\textwidth]{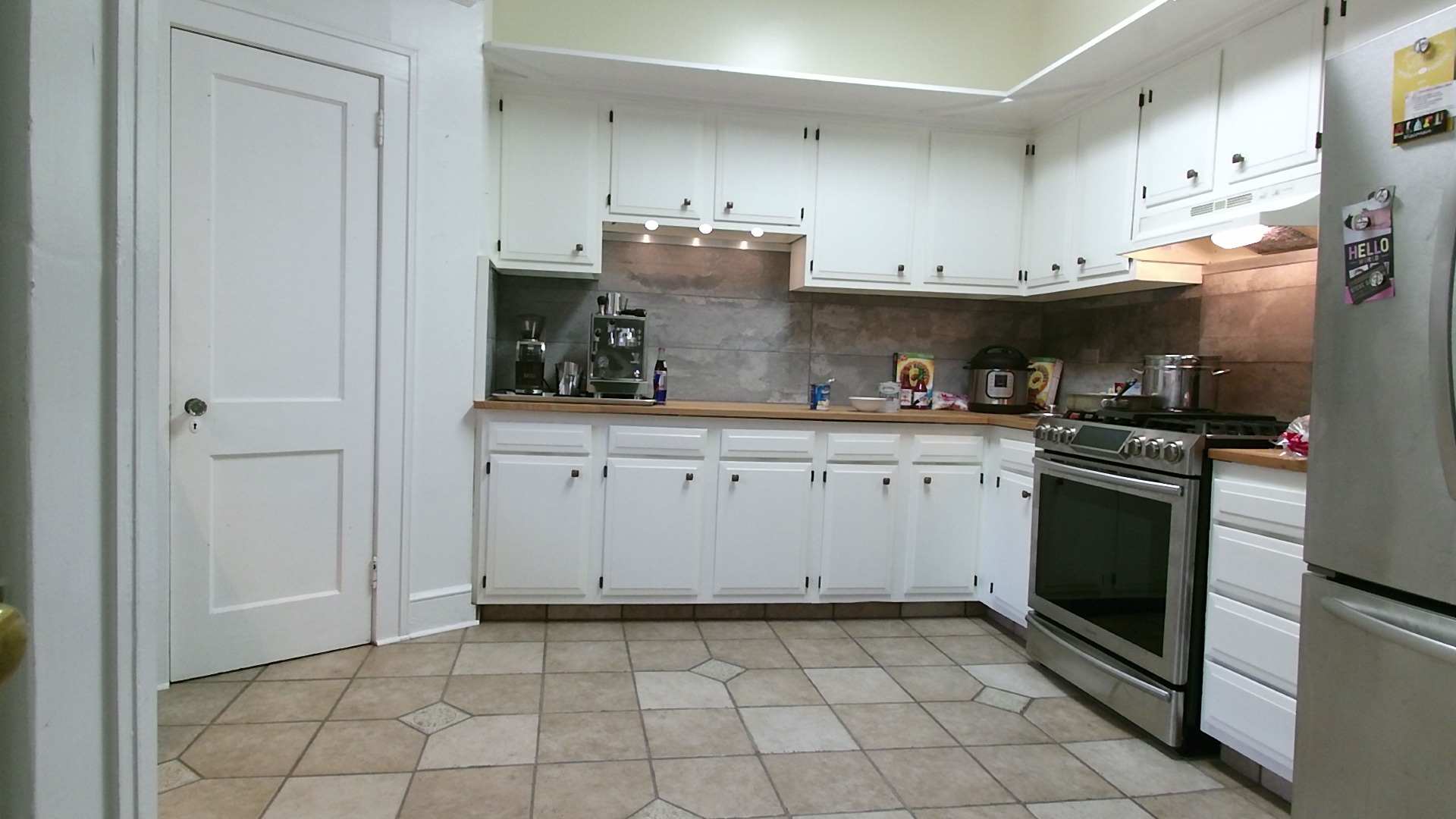}&
		\includegraphics[width=0.23\textwidth]{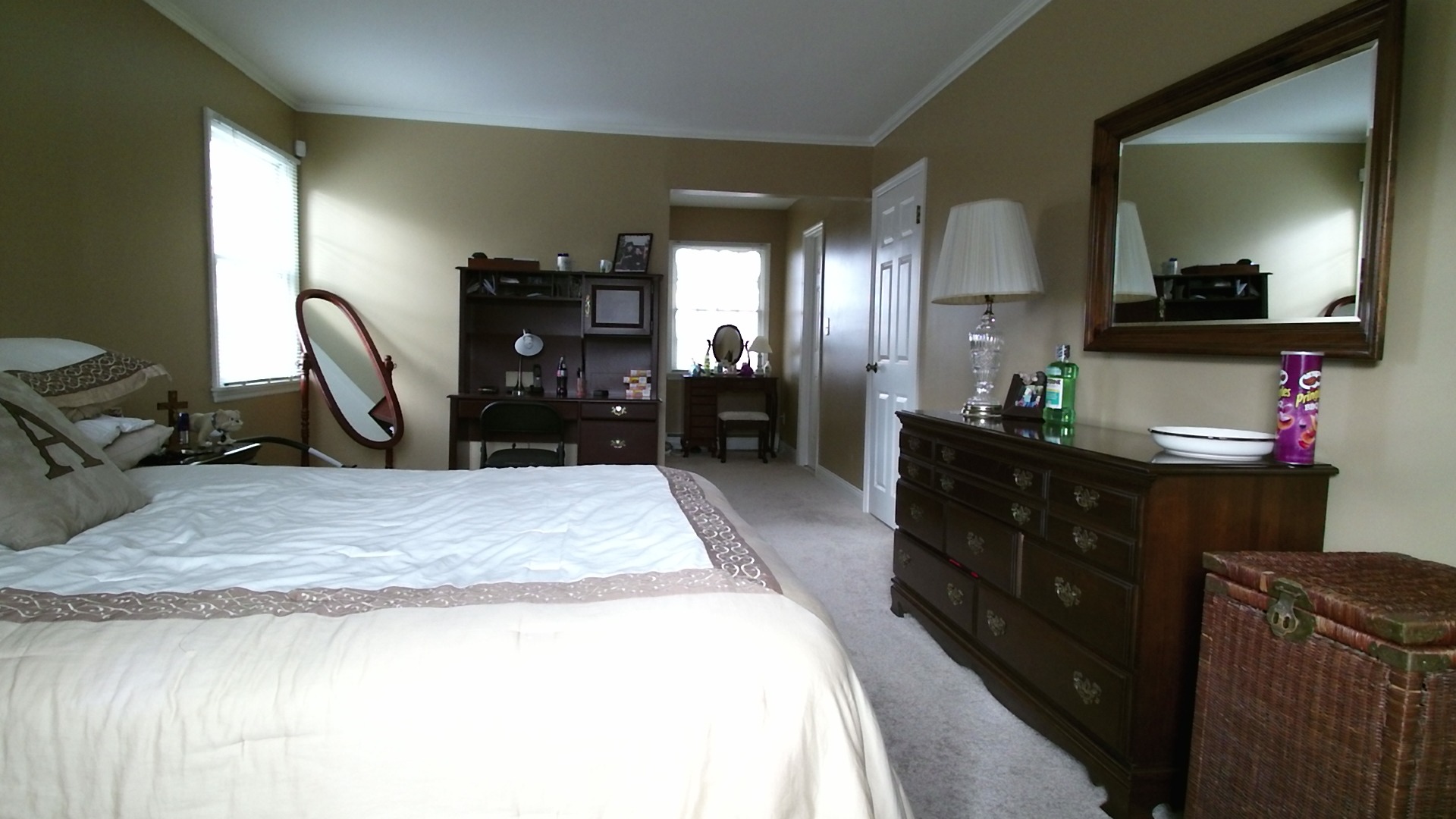}&
		\includegraphics[width=0.23\textwidth]{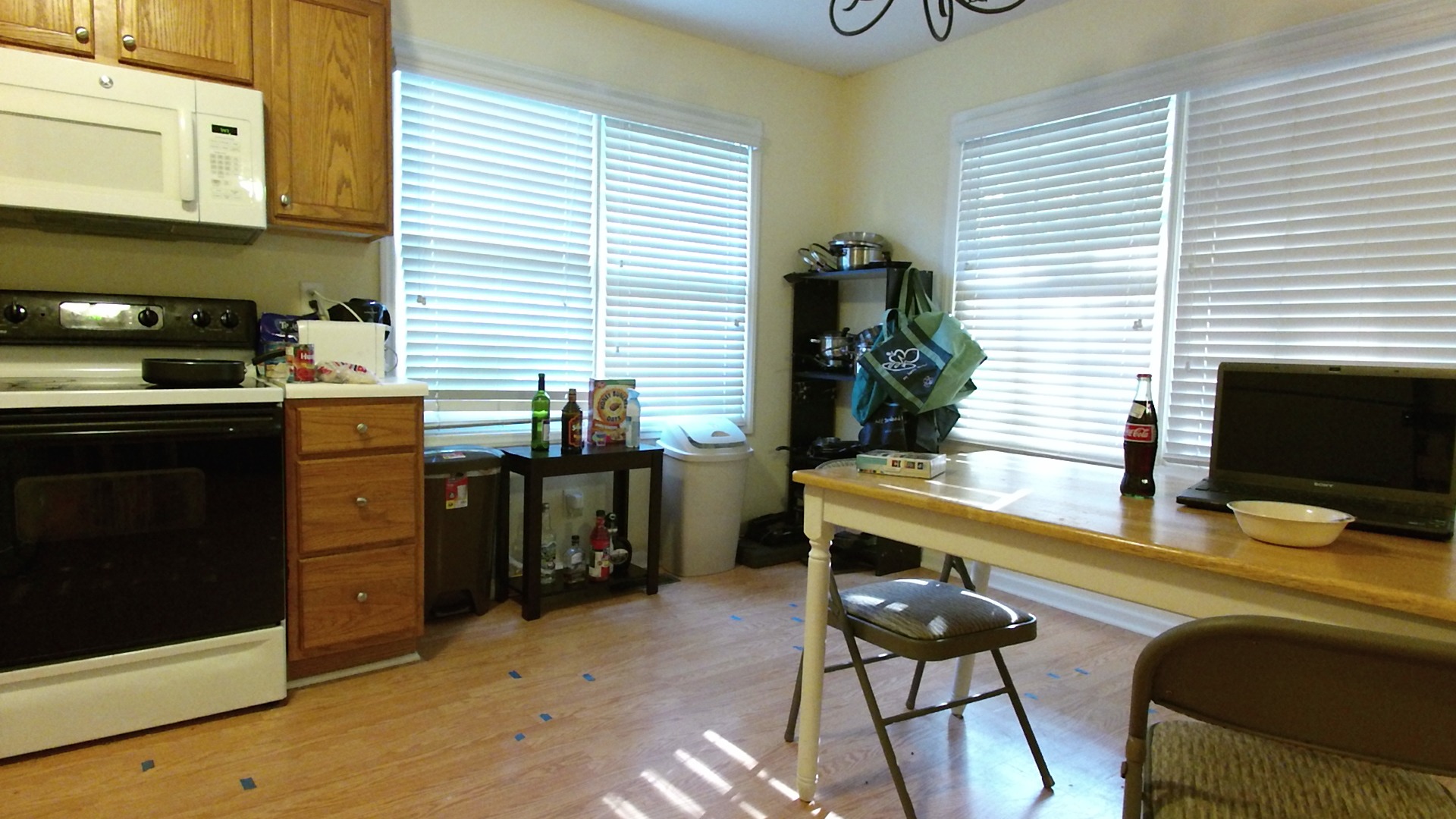}&
		\includegraphics[width=0.23\textwidth]{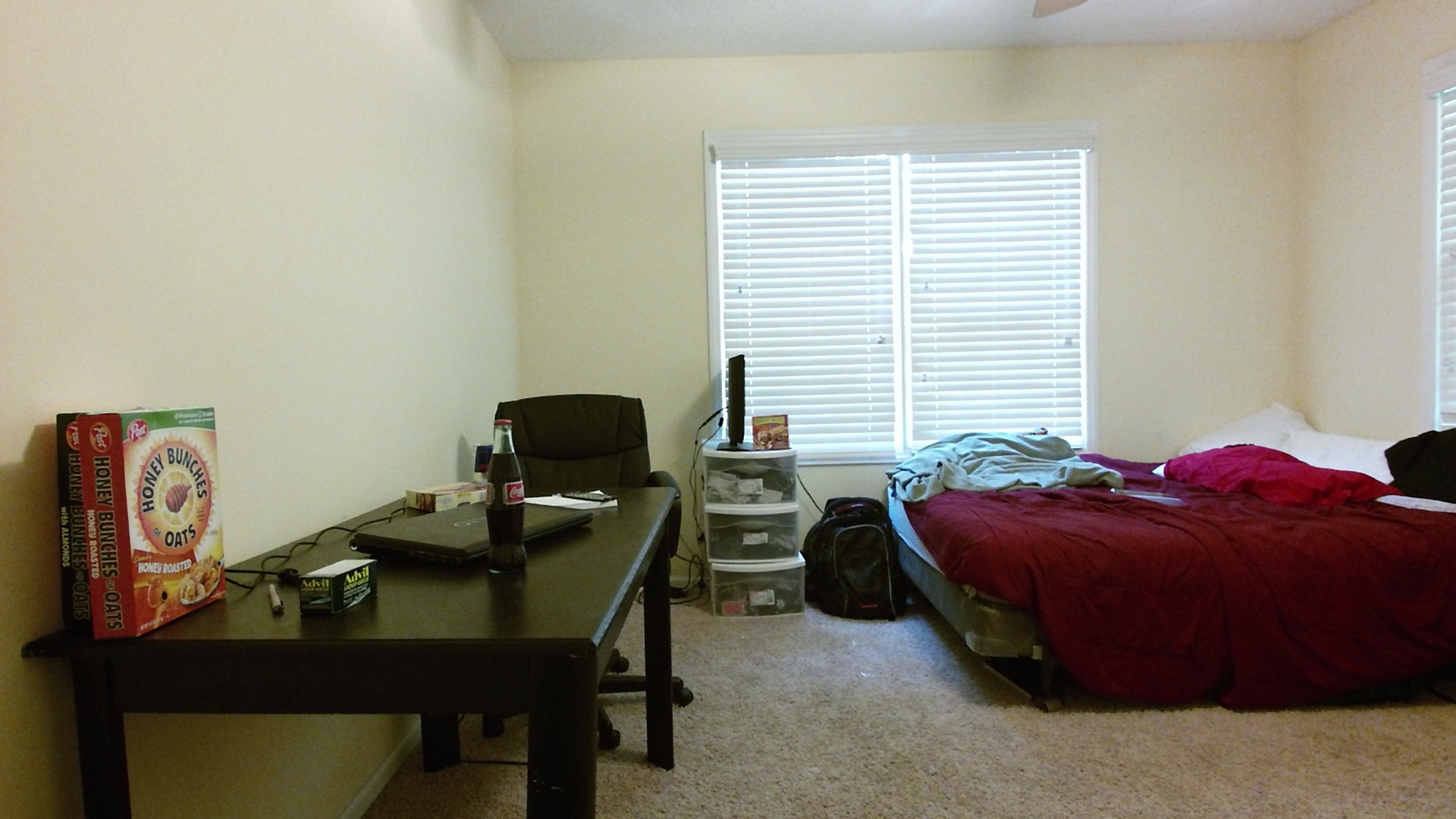}\\
		(a) Easy: Home\_005\_2 & (b) Easy: Home\_015\_1 & (c) Medium: Home\_001\_2 & (d) Medium: Home\_014\_2\\
		\includegraphics[width=0.23\textwidth]{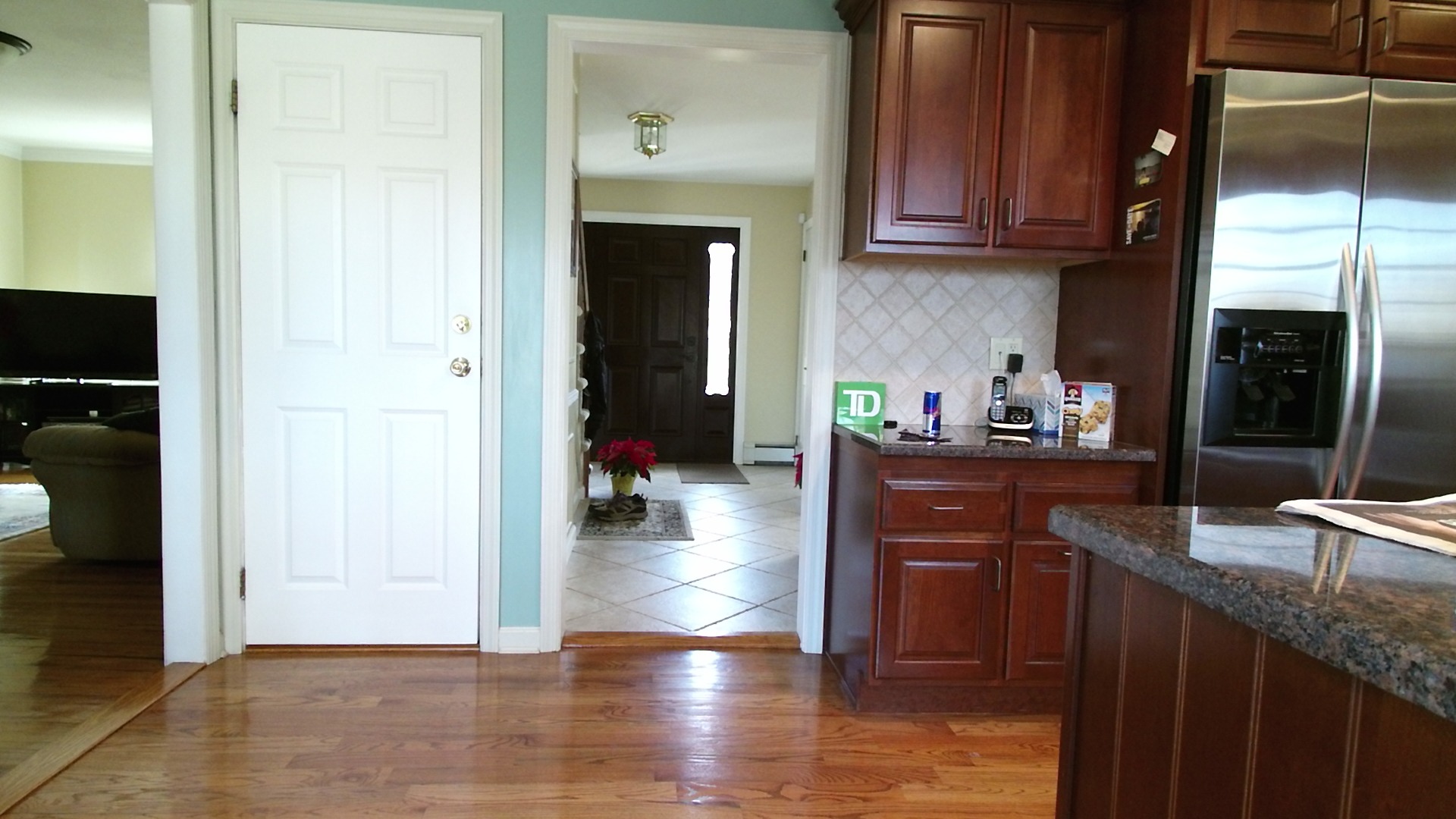}&
		\includegraphics[width=0.23\textwidth]{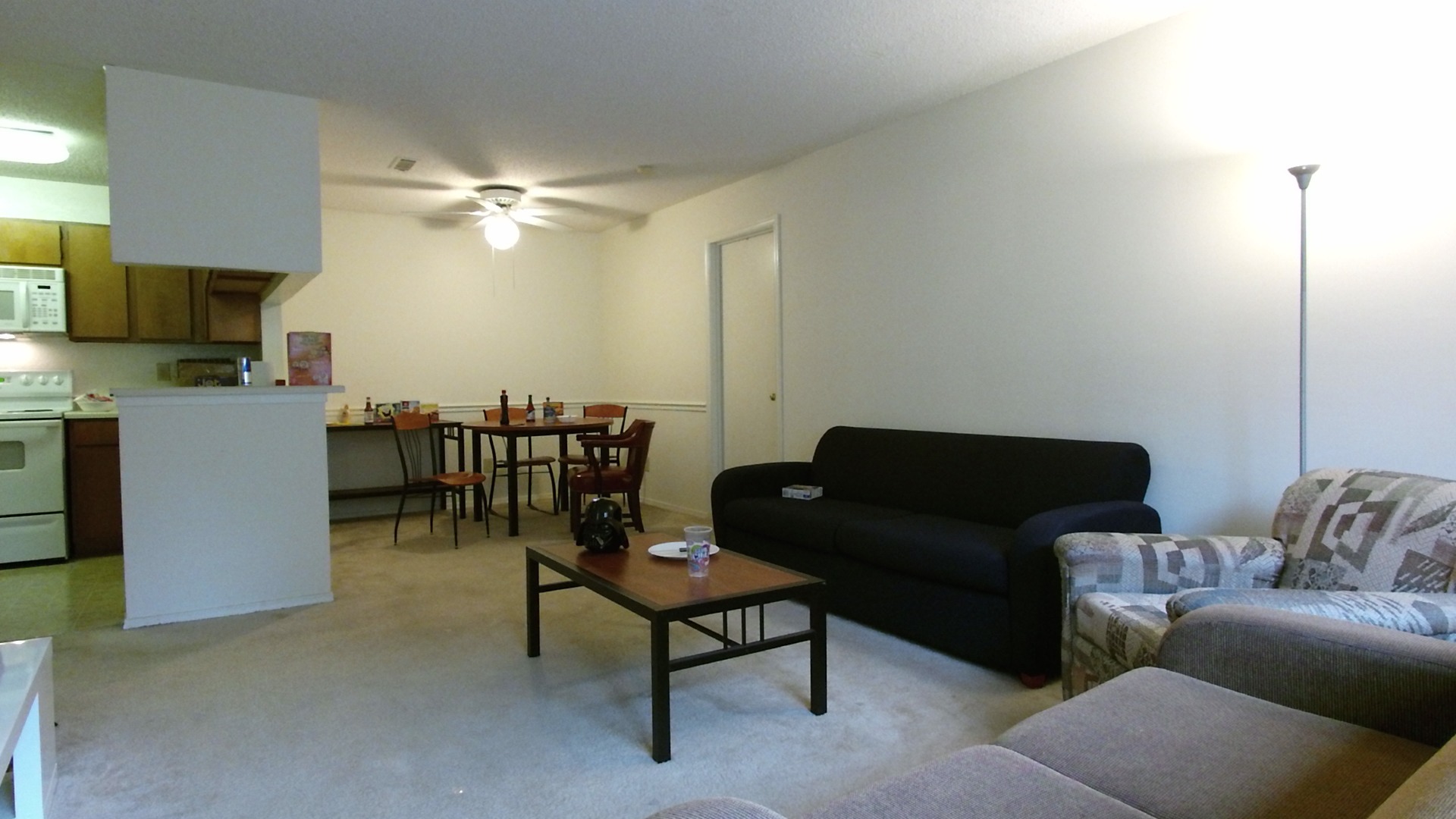}&
		\includegraphics[width=0.23\textwidth]{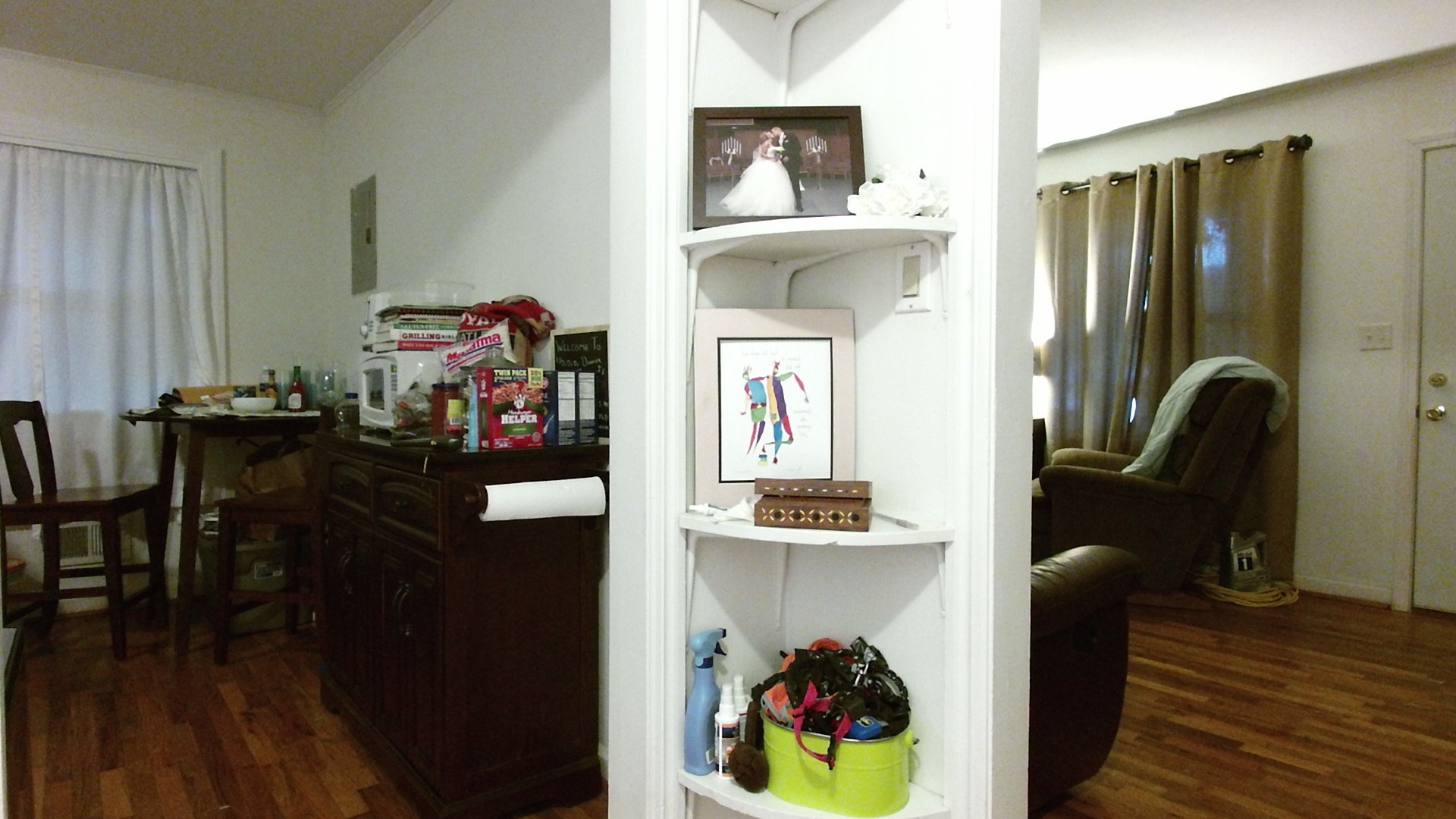}&
		\includegraphics[width=0.23\textwidth]{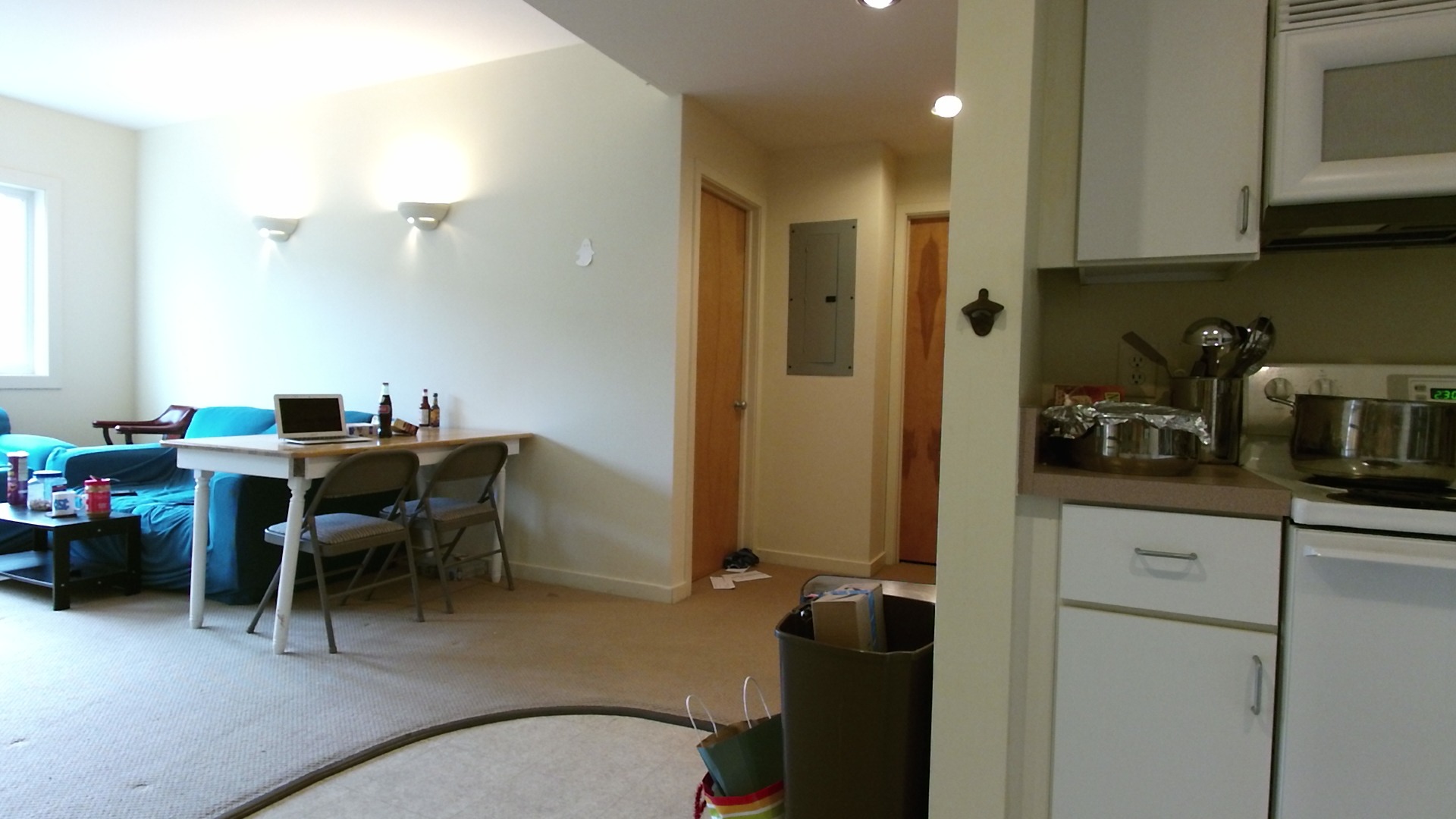}\\
		(e) Medium: Home\_016\_1 & (f) Hard: Home\_003\_2 & (g) Hard: Home\_004\_2 & (h) Hard: Home\_013\_1\\
   	\end{tabular}
\end{center}
\vspace{-0.3cm}
\caption{ Test scenes from AVBD of 3 different difficulty levels (as in \cite{schmid2019iros}), where we define a simple environment consisting of a single small room, a medium difficult apartment with a large room or with an additional small room, e.g., a bathroom or open space, and finally a hard apartment with multiple large rooms. Easy: Home\_005\_2 and Home\_015\_1; Medium: Home\_001\_2, Home\_016\_1 and Home\_014\_2, and Hard: Home\_003\_2, Home\_004\_2 and Home\_013\_1.}
\label{fig:exp_scenes}
\end{figure*}

We tested our approach on the Active Vision Dataset Benchmark~\cite{ammirato2017dataset}, a public benchmark for active visual search that contains more than 30,000 RGBD images taken in 15 different indoor environments and 33 different target objects. 
Consistently with~\cite{schmid2019iros}, we classify each scene in the dataset as simple, medium, or hard for the visual search task, where we define a simple environment consisting of a single small room, a medium difficult apartment with a large room or with an additional small room --\eg a bathroom or an open space--, and finally a hard apartment with multiple large rooms. 
We use in our experiments two simple (Home\_005\_2 and Home\_015\_1), three medium (Home\_001\_2, Home\_016\_1, Home\_014\_2), and three hard apartments (Home\_003\_2, Home\_004\_2, Home\_013\_1).
Some examples of these scenarios are shown in Fig.~\ref{fig:home_maps} and Fig.~\ref{fig:exp_scenes}.

\begin{table*}[t!]
\caption{Results on three scenes from AVDB using GT objects annotations. All methods are compared using the protocol defined in~\cite{schmid2019iros}. The asterisk ${}^{(*)}$ indicates that the evaluation is performed on a different subset of objects.}
\resizebox{\textwidth}{!}{
\begin{tabular}{|c|c|c|c|c|c|c|c|c|c|c|c|c|}
\hline
\multirow{2}{*}{} & \multicolumn{3}{c|}{Easy (Home\_005\_2)} & \multicolumn{3}{c|}{Medium (Home\_001\_2)} & \multicolumn{3}{c|}{Hard (Home\_003\_2)} & \multicolumn{3}{c|}{Avg.} \\ \cline{2-13} 
 & SR $\uparrow$ & APL $\downarrow$ & SPL $\uparrow$ & SR $\uparrow$ & APL $\downarrow$ & SPL $\uparrow$ & SR $\uparrow$ & APL $\downarrow$ & SPL $\uparrow$ & SR $\uparrow$ & APL $\downarrow$ & SPL $\uparrow$ \\ \hline
Random Walk & 0.32 & 74.00 & 0.19  & 0.11 & 74.48  & 0.21   & 0.10 & 79.27 & 0.17  & 0.18 & 75.91 & 0.19   \\ \hline
EAT~\cite{schmid2019iros}  & 0.77 & 12.20 & - & 0.73 & 16.20 & - & 0.58 & 22.10 & - & 0.69 & 16.80 & - \\ \hline
DQN${}^{(*)}$~\cite{deep_q_n_AVDB}
& \textit{1.00} & \textit{11.06} & - & \textit{0.69}  & \textit{18.15}  & - & - & -  &  -  & - & - & - \\ \hline
DQN-TAM${}^{(*)}$~\cite{deep_q_n_AVDB_2}
& \textit{0.98} & \textit{17.85} & - & \textit{0.60}  & \textit{24.19}  & - & - & -  &  -  & - & - & - \\ \hline
POMP~\cite{pomp2020bmvc} & 0.98 & 13.60 & 0.72 & 0.73 & 17.10 & 0.80 & 0.56 & 20.5 & 0.72 & 0.76 & 17.07 & 0.75 \\ \hline
\methname & 0.98 & 11.93 & 0.74 & 0.80 & 17.86 & 0.76 & 0.92 & 24.52 & 0.66 & \textbf{0.90} & 18.10 &  0.72\\ \hline
\end{tabular}}
\label{table:comp_avd_eat}
\end{table*}

We consider three metrics: \emph{Success Rate} (SR)~\cite{on_evaluation_of_embodied} defined as the percentage of times the agent successfully reaches one of the destination poses (as provided in AVDB) over the total number of trials (a larger value indicates a more effective search); \emph{Average Path Length} (APL) defined as the total number of poses visited by the agent, among the successful episodes, divided by the number of successful episodes (a lower value indicates a higher absolute efficiency); and \emph{Success weighted by Path Length} (SPL)~\cite{on_evaluation_of_embodied} defined as: 
\begin{equation}
    SPL = \frac{1}{N}\sum_{i=1}^{N}S_i\frac{l_i}{\max(p_i,l_i)},
\end{equation}
where $l_i$ is the length of the shortest path between the goal and the target for an episode, $p_i$ is the length of the path taken by an agent in an episode and $S_i$ is a binary indicator of success in episode $i$ (a larger value indicates a higher absolute efficiency). An episode is considered successful if the agent reaches the destination pose given by AVDB in a fixed number of steps (200 in our experiments).
It should be noted that the SPL metric is occasionally referred to as ASPPL in some works~\cite{ammirato2018target}.

\subsection{Quantitative results}
\label{sec:exp:aquantitative}

We compare our proposed approach \methname against a random walk baseline --\ie we allow the agent to randomly select an action among all the feasible ones at each time step--, and four state-of-the-art approaches, namely: EAT~\cite{schmid2019iros}, DQN~\cite{deep_q_n_AVDB}, DQN-TAM~\cite{deep_q_n_AVDB_2}, and our previous work POMP~\cite{pomp2020bmvc}.
The latter is the only unsupervised method, while the former three need training data to learn the policy.
Since no official code for published methods is available, we are only able to compare results with them following the protocol proposed in~\cite{schmid2019iros}. With respect to the standard protocol defined in the benchmark paper~\cite{ammirato2017dataset}, this protocol provides results only using GT annotations for object detection, and on a limited number of scenes.
Moreover, it uses only a subset of target objects.
DQN and DQN-TAM use only two scenes (one easy and one medium), thus the average column is not meaningful for a fair comparison.
Additionally, DQN and DQN-TAM use a different subset of objects in their evaluation.

From results reported in Table~\ref{table:comp_avd_eat} we can clearly see that our approach \methname outperforms EAT in terms of SR, with a little increment in APL, which is reasonable since we are now considering more challenging situations, as we will deeply explore in Sect.~\ref{sec:exp:aqualitative}. As for the comparison between \methname and DQN, we note that the DQN approach outperforms our method in the easy scenario, but in the medium case we outperform the competitor in SR with a comparable APL. 

It is worth noting that for achieving these results both DQN approaches require 13 scenarios for training the best policy, while our method requires no training at all.

Results using the object detector provided by \cite{ammirato2018target} are reported in Table~\ref{table:ablation_using_detector} for POMP and \methname.
Again, we can appreciate a strong increment in SR of $35\%$ with a limited decrease in SPL of $0.03$, mostly due to the ability of our proposed method to handle more complex cases.

\begin{figure*}[]
\begin{center}
	\begin{tabular}{@{}c@{}c@{}}
		\includegraphics[width=0.5\textwidth]{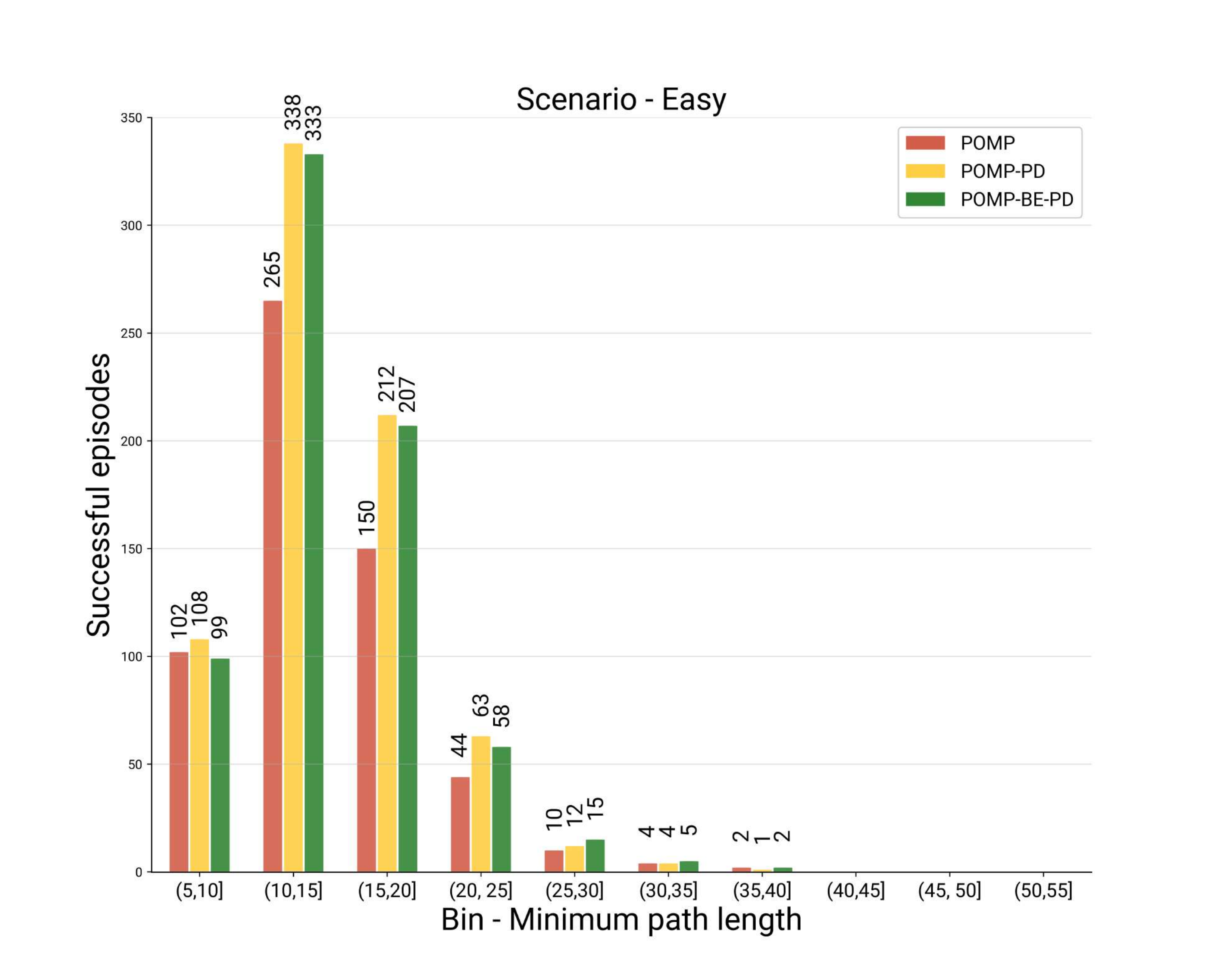}&
             \includegraphics[width=0.5\textwidth]{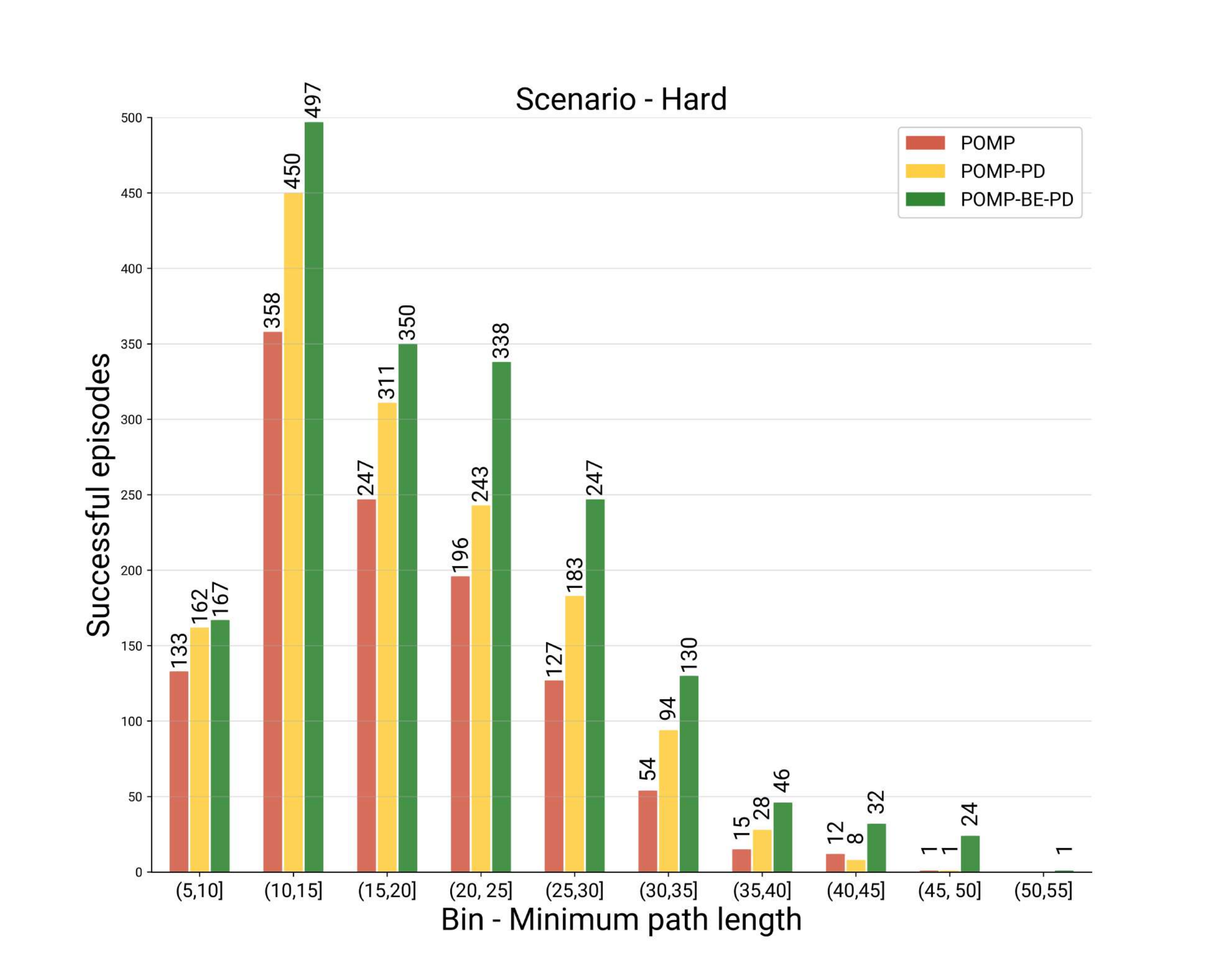} \\
             (a) Easy & (b) Hard \\
             \includegraphics[width=0.5\textwidth]{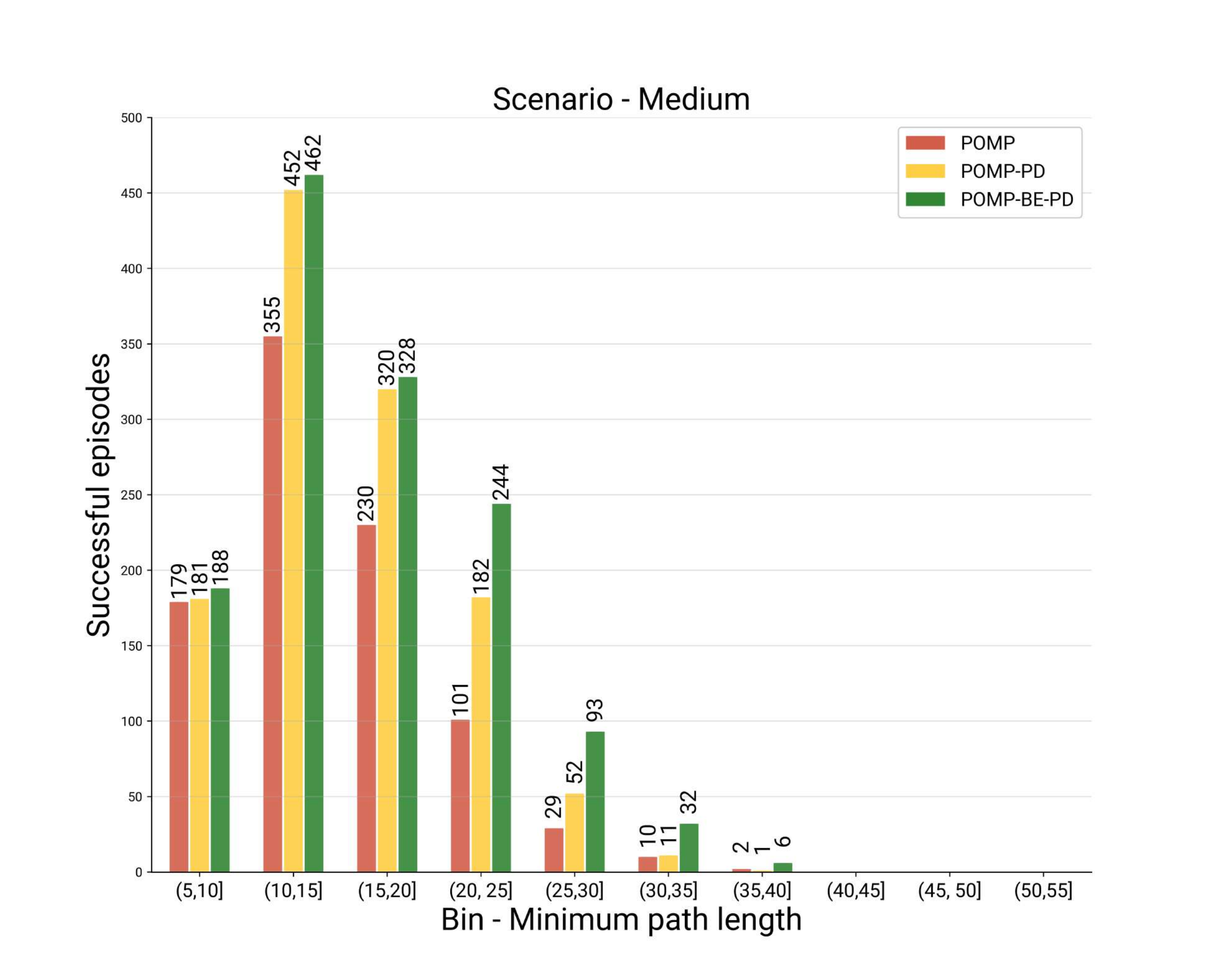} &
		\includegraphics[width=0.5\textwidth]{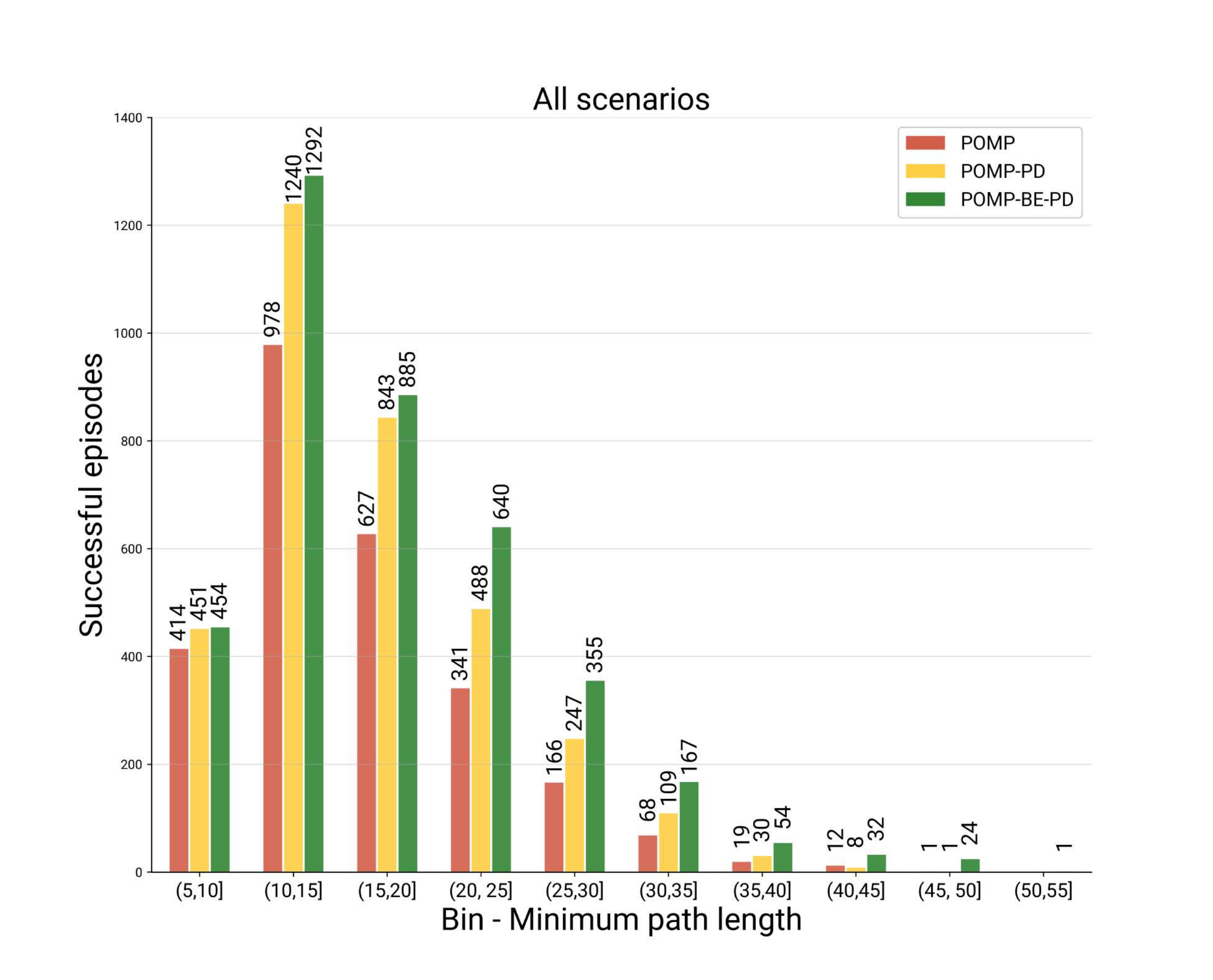}\\ 
  \\ (c) Medium & (d) All
   	\end{tabular}
\end{center}

\vspace{-0.3cm}
\caption{We aggregated the episodes by the minimum number of steps to reach the object, thus incorporating the difficulty of the episode. 
In figure (a) the results for the Easy scenarios: Home\_005\_2 and Home\_015\_1; in figure (b) Hard Scenarios: Home\_003\_2, Home\_004\_2, Home\_013\_1; in figure (c) the Medium one: Home\_001\_2, Home\_014\_2 and Home\_016\_1; figure (d) the sum on all scenarios. Results using the object detector provided by \cite{ammirato2018target}, both during planning and docking. }

\label{fig:path_length_different_planner}
\extralabel{fig:path_length_different_planner:a}{(a)}
\extralabel{fig:path_length_different_planner:b}{(b)}
\extralabel{fig:path_length_different_planner:c}{(c)}
\extralabel{fig:path_length_different_planner:d}{(d)}
\end{figure*}

\begin{figure}[ht]
\caption{Percentage of error of POMP, POMP-PD and \methname. The errors are categorised into three types: Localisation, Docking and Other. We used the object detector provided by~\cite{ammirato2018target}, during both planning and docking.}
\centering
\includegraphics[width=0.5\textwidth]{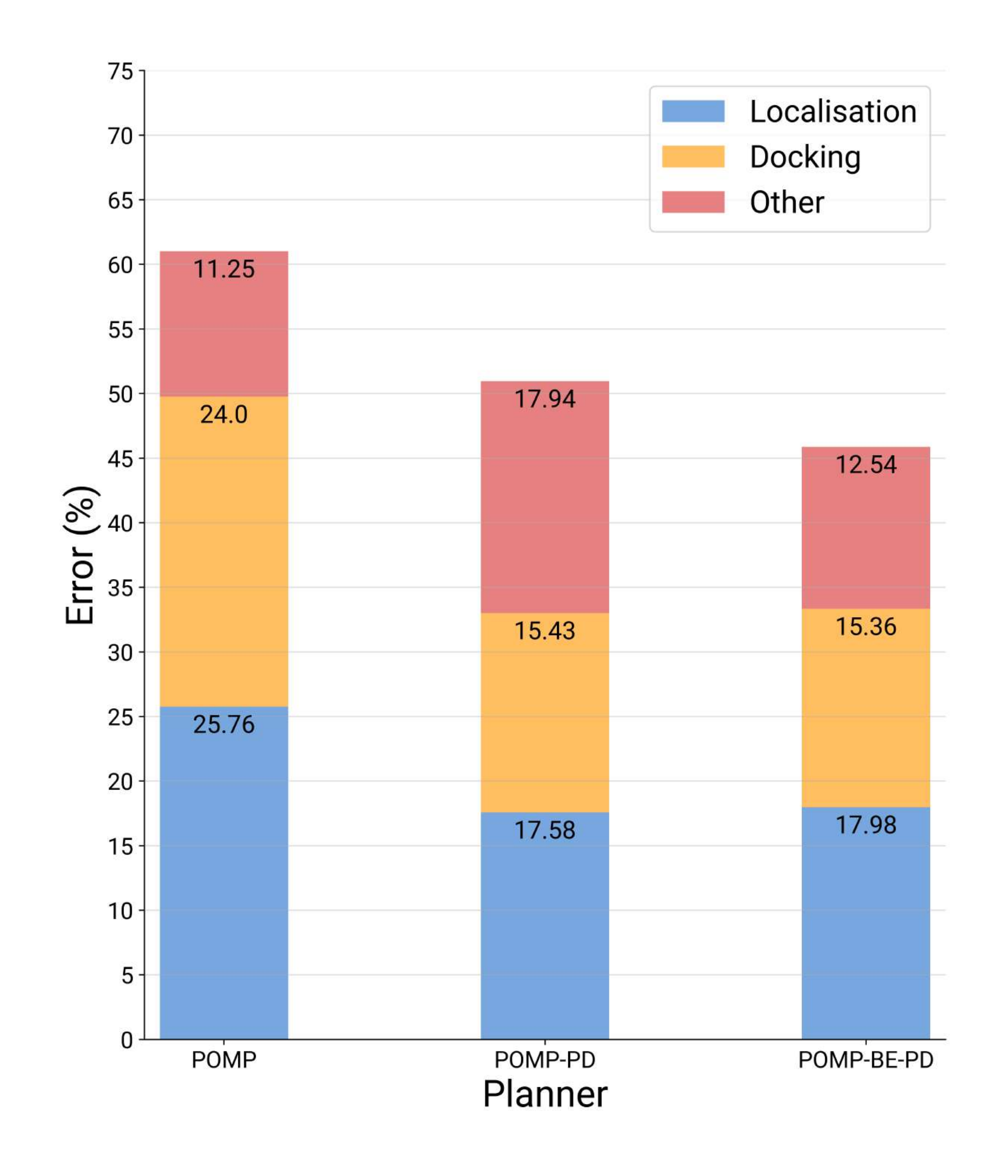}
\label{fig:error_by_type_all_home}
\end{figure}


\subsection{Ablation studies}
\label{sec:exp:ablation}

\begin{table*}[t!]
	\caption{Result of different versions of improved POMP with more scenes per difficulty level in AVD. POMP-BE is POMP with modification of the belief update. Result using the ground truth annotations instead of the detector and with $2^{10}$ simulations.}
    \centering
	\resizebox{.6\textwidth}{!}{
		\begin{tabular}{|c|c|c|c|c|c|c|c|}
			  \hline
			\multirow{2}{*}{Difficulty} &
			\multirow{2}{*}{Scene} &
			\multicolumn{3}{c|}{POMP\cite{pomp2020bmvc}} &
			\multicolumn{3}{c|}{POMP-BE}  \\ \cline{3-8} 
			                        &              & SR $\uparrow$ & APL $\downarrow$ & SPL $\uparrow$ & SR $\uparrow$ & APL $\downarrow$ & SPL $\uparrow$ \\ \hlineB{1.8}
			\multirow{3}{*}{Easy}   & Home\_005\_2 &  0.94  & 12.96  &  0.78           & 0.93 & 12.26 & 0.78  \\ \cline{2-8} 
			                        & Home\_015\_1 &  0.75  & 23.66  &  0.61           & 0.73 & 17.04 & 0.72  \\ \cline{2-8} 
			                        & Avg.         & \textbf{0.84}   & 18.31  &  0.70            & 0.83 & \textbf{14.65} & \textbf{0.75}  \\ \hlineB{1.8}
			\multirow{4}{*}{Medium} & Home\_001\_2 & 0.80   & 18.20   &  0.72           & 0.81 & 19.95 & 0.69  \\ \cline{2-8} 
			                        & Home\_014\_2 & 0.76   & 41.07  &  0.51           & 0.90 & 19.99 & 0.66   \\ \cline{2-8} 
			                        & Home\_016\_1 & 0.71   & 29.64  &  0.56           & 0.83 & 36.55 & 0.60  \\ \cline{2-8} 
			                        & Avg.         & 0.76   &29.64   & 0.60             & \textbf{0.85} & \textbf{25.50} & \textbf{0.65}  \\ \hlineB{1.8}
			\multirow{4}{*}{Hard}   & Home\_003\_2 & 0.43   & 21.90 & 0.65              & 0.79 & 31.93 & 0.58  \\ \cline{2-8} 
			                        & Home\_004\_2 & 0.45   & 66.20 & 0.39             & 0.57 & 47.71 & 0.49  \\ \cline{2-8} 
			                        & Home\_013\_1 & 0.55   & 49.72 & 0.50             & 0.74 & 53.11 & 0.56 \\ \cline{2-8} 
			                        & Avg.         & 0.48   & 45.94 & 0.51             & \textbf{0.70}  & \textbf{44.25} & \textbf{0.54}  \\ \hlineB{1.8}
			\multicolumn{2}{|c|}{Average}          & 0.67   & 32.92 & 0.59             & \textbf{0.79}  & \textbf{29.82} & \textbf{0.63}   \\ \hline
		\end{tabular}}
	\label{table:ablation_pomp_pomb-b}
\end{table*}

\begin{table*}[t!]
\caption{Results of POMP and variations of \methname with more scenes per difficulty level in AVD using the real detector provided by~\cite{ammirato2018target}.}
\resizebox{\textwidth}{!}{
\begin{tabular}{|c|c|c|c|c|c|c|c|c|c|c|c|c|c|}
\hline
\multirow{2}{*}{Difficulty} &
  \multirow{2}{*}{Scene} &
  \multicolumn{3}{c|}{POMP\cite{pomp2020bmvc}} &
  \multicolumn{3}{c|}{POMP-BE} &
  \multicolumn{3}{c|}{POMP-PD} &
  \multicolumn{3}{c|}{\methname (Proposed)} \\ \cline{3-14} 
                        &               & SR $\uparrow$ & APL $\downarrow$ & SPL $\uparrow$ & SR $\uparrow $ & APL $\downarrow $ & SPL $\uparrow$ & SR $\uparrow $ & APL $\downarrow $ & SPL $\uparrow$ & SR $\uparrow$ & APL $\downarrow$ & SPL $\uparrow$ \\ \hlineB{1.8}
\multirow{3}{*}{Easy}   & Home\_005\_2  & 0.60   & 17.90    & 0.68 & 0.58 & 16.18 & 0.71  & 0.81   & 26.08    & 0.51    & 0.79  & 22.70  & 0.56 \\ \cline{2-14} 
                        & Home\_015\_1  & 0.49  & 34.76   & 0.46 & 0.45 & 38.76 & 0.51  & 0.55   & 35.34    &  0.42    & 0.54 & 30.50  & 0.49 \\ \cline{2-14} 
                        & Avg.          & 0.54  & \textbf{26.33}   & 0.57 & 0.52 & 27.47 & \textbf{0.61}   & \textbf{0.68}   & 30.71    &  0.47    & 0.67 & 26.60  & 0.53 \\ \hlineB{1.9}
\multirow{4}{*}{Medium} & Home\_001\_2  & 0.40   & 20.73   & 0.62 & 0.39 & 19.36 & 0.63  & 0.50    & 31.00    &  0.47    & 0.57 & 28.50  & 0.53 \\ \cline{2-14} 
                        & Home\_014\_2  & 0.53  & 47.60   & 0.48 & 0.60 & 18.52 & 0.64  & 0.60    & 45.79    &  0.41    & 0.66 & 21.38  & 0.56 \\ \cline{2-14} 
                        & Home\_016\_1  & 0.29  & 50.23   & 0.42 & 0.28 & 47.05 & 0.45  & 0.36   & 57.73    &  0.33    & 0.41 & 53.26  & 0.39 \\ \cline{2-14} 
                        & Avg.          & 0.41  & 39.52   & \textbf{0.51} & 0.42 & \textbf{28.31} & 0.57  & 0.49   & 44.84    &  0.40     & \textbf{0.55} & 34.38  & 0.49 \\ \hlineB{1.9}
\multirow{4}{*}{Hard}   & Home\_003\_2  & 0.19  & 26.60    & 0.53 & 0.33 & 30.53 & 0.55  & 0.39   & 62.36    &  0.34    & 0.48 & 42.86  & 0.42 \\ \cline{2-14} 
                        & Home\_004\_2  & 0.42  & 69.84   & 0.36 & 0.55 & 47.31 & 0.47  & 0.44   & 70.26    & 0.31     & 0.54 & 61.93  & 0.36 \\ \cline{2-14} 
                        & Home\_013\_1  & 0.25  & 61.41   & 0.46 & 0.31 & 77.09 & 0.44  & 0.26   & 62.80    & 0.37     & 0.34 & 54.38  & 0.46 \\ \cline{2-14} 
                        & Avg.          & 0.29  & 52.62   & \textbf{0.45} & 0.40 & \textbf{51.64} & 0.49  & 0.36   & 65.14    & 0.34     & \textbf{0.45} & 53.06  & 0.41 \\ \hlineB{1.9}
\multicolumn{2}{|c|}{Average}           & 0.40  & 41.13  & \textbf{0.50} & 0.44 & \textbf{38.85} & 0.55   & 0.49   & 48.92    & 0.40      & \textbf{0.54} & 39.44  & 0.47  \\ \hline
\end{tabular}}
\label{table:ablation_using_detector}
\end{table*}

We provide also an ablation study to deeply analyse the contributions of the different terms in our model. In the following we will answer some questions by comparing the proposed method with some partial versions of it  considering only the new belief update (called POMP-BE) and considering only the probabilistic detection (called POMP-PD).

\vspace{.5em}
\noindent\textbf{Belief update:}
\textit{Does the new belief update reduce the episode length? What are the benefits of the new belief update when navigating difficult scenarios?}\\
In Fig.~\ref{fig:path_length_different_planner} we aggregate the episodes by their difficulty, grouped by the minimum path length for the episode to reach the target. In Fig.~\ref{fig:path_length_different_planner:a} we aggregated the results for the easy scenario (Home\_005\_2 and Home\_015\_1); in Fig.~\ref{fig:path_length_different_planner:b} for the hard scenario (Home\_003\_2, Home\_004\_2, Home\_013\_1); in Fig.~\ref{fig:path_length_different_planner:c} for the medium one (Home\_001\_2, Home\_014\_2 and Home\_016\_1); finally for Fig.~\ref{fig:path_length_different_planner:d} we aggregated the results for all the scenario present in AVDB. 
From this reports we can derive that, by removing from the belief update locations already observed, we can optimise the exploration phase and thus reduce the episode length.

Furthermore, Table~\ref{table:ablation_pomp_pomb-b} analyzes the impact of the new belief update isolating all the possible causes of error, \ie we swap the object detector with the ground truth annotations eliminating the source of false positive and miss detection both during the planning and docking phases.
In the easy scenario, we maintain the same SR while reducing the APL and increasing the efficiency (\ie SPL). We hypothesise that the simple layout (Fig.~\ref{fig:home_maps:a}, \ref{fig:home_maps:b}) and the concentration of minimum path length (Fig.~\ref{fig:path_length_different_planner:a}) do not allow the belief update to be beneficial.

However, the more difficult the scenario, with more possible object locations and complicated spatial layouts, the higher the improvement in performance.
Starting from the medium scenario, we increase the SR from 0.76 to 0.85 while decreasing the APL from 29.64 to 25.50, with an increment of the total efficiency from 0.6 to 0.65. A similar result can be seen for the hard scenario.

\vspace{.5em}
\noindent\textbf{Probabilistic Detection \& Docking:}

\textit{Does Probabilistic Detection reduce the number of false positives? Is there a way to improve the docking, also considering the knowledge gathered during the planning?}\\
To answer the first question, we analyze the different types of failure that can occur during the episodes.
More specifically, we consider three types of error:
\textit{Localisation}, if during POMCP exploration the exit condition is verified, but the target object is not actually present in the FOV.
\textit{Docking}, if in the last pose of the POMCP exploration the agent correctly detects the target, but it fails to reach the destination pose defined by AVDB;
\textit{Other}, \eg if the agent is unable to detect the target object within the time limit, or the agent performs action not allowed during the path;

In Fig.~\ref{fig:error_by_type_all_home} we provide the percentage of error for each planner, averaged over all scenarios. First of all, we can notice a significant reduction ($\sim$32\% decrease) of false positives when introducing the Probabilistic Detection approach, thus increasing the robustness of our method. 
Moreover, using the knowledge gathered during the planning provides a reliable mechanism to increase the robustness during the docking phase. Indeed, if we look at Fig.~\ref{fig:error_by_type_all_home} we can appreciate a $\sim$35.7\% decrease of \textit{Docking} error.

To measure the impact of Probabilistic Detection, in table \ref{table:ablation_using_detector} we conduct an ablation study isolating the Probabilistic approach from the belief update, both using the detector during the planning and docking. In all the scenarios, POMP-PD increments the SR by a large margin ($19 \%$ - $25\%$) over our previous formulation POMP. However, we note also an increment of the APL and a reduction of the total efficiency (SPL). 

This is not surprising: indeed, if the agent needs to be more confident and robust against false positives, it must require more steps to increase the probability of the target location, and bring that to be $\geq$ the threshold $\tau$.

\subsection{Qualitative results}
\label{sec:exp:aqualitative}

In Fig.~\ref{fig:step_by_step} we visualise an episode by our proposed approach, in which it is possible to appreciate the evolution of the probability distribution over the locations and the robustness to a false positive. The starting pose is defined in Fig.~\ref{fig:step_by_step:a}. From Fig.~\ref{fig:step_by_step:b} to \ref{fig:step_by_step:c} the agent explores the top part of the environment without success. In Fig.~\ref{fig:step_by_step:d} the robot encounters a false positive: the update rule defined in Eq.~\ref{eqn:update_rule} with the generated probability map are providing a robust framework for not stopping the exploration. Indeed, the standard POMP defined in \cite{pomp2020bmvc}, in the same situation, would have stopped the episode. 

Moreover, we can note that in the area unexplored by the agent (the right part of the environment) we are raising the probabilities: if we do not locate the target elsewhere, the object must be in this area.
Finally, in Fig.~\ref{fig:step_by_step:e} we locate a zone with a high probability of containing the target, and in Fig.~\ref{fig:step_by_step:f} we locate the searched object.

\section{Conclusion}
\label{sec:conclusion}
In this paper we presented \methname, our proposed approach to solve Active Visual Search (AVS) in known environments. Based on a POMCP planner, \methname learns the policy online by efficiently exploiting the information of the 2D floor map of the environment; as a consequence, our method does not require any expensive training, both in time and computational resources. 
To cope with imperfect object detectors, with high number of false positives and miss-detection that could have a dramatic effect on the overall success rate, we transitioned from a deterministic detection to a \textit{Probabilistic} one.
After every step in the real world, a Bayesian inference, combined with a probability distribution over all possible object locations, allows us to reduce the false positives error by $32\%$. Consequently, to handle the restricted \textit{belief space} of the original POMCP in the AVS domain, we introduce a new belief update considering, at each time step, all the possible positions that have not been observed yet. 
We evaluate extensively our method, following the AVDB benchmark, achieving state-of-the-art results. On top of that, with several ablation studies, we demonstrated the strength of our method. On average over all the environments, we increase the success rate by a significant $14\%$ while decreasing the average path length by $4\%$ with respect to our previous formulation POMP.

\bibliographystyle{IEEEtran}
\bibliography{ref}

\begin{IEEEbiography}[{\includegraphics[width=1in,height=1.2in,clip,keepaspectratio]{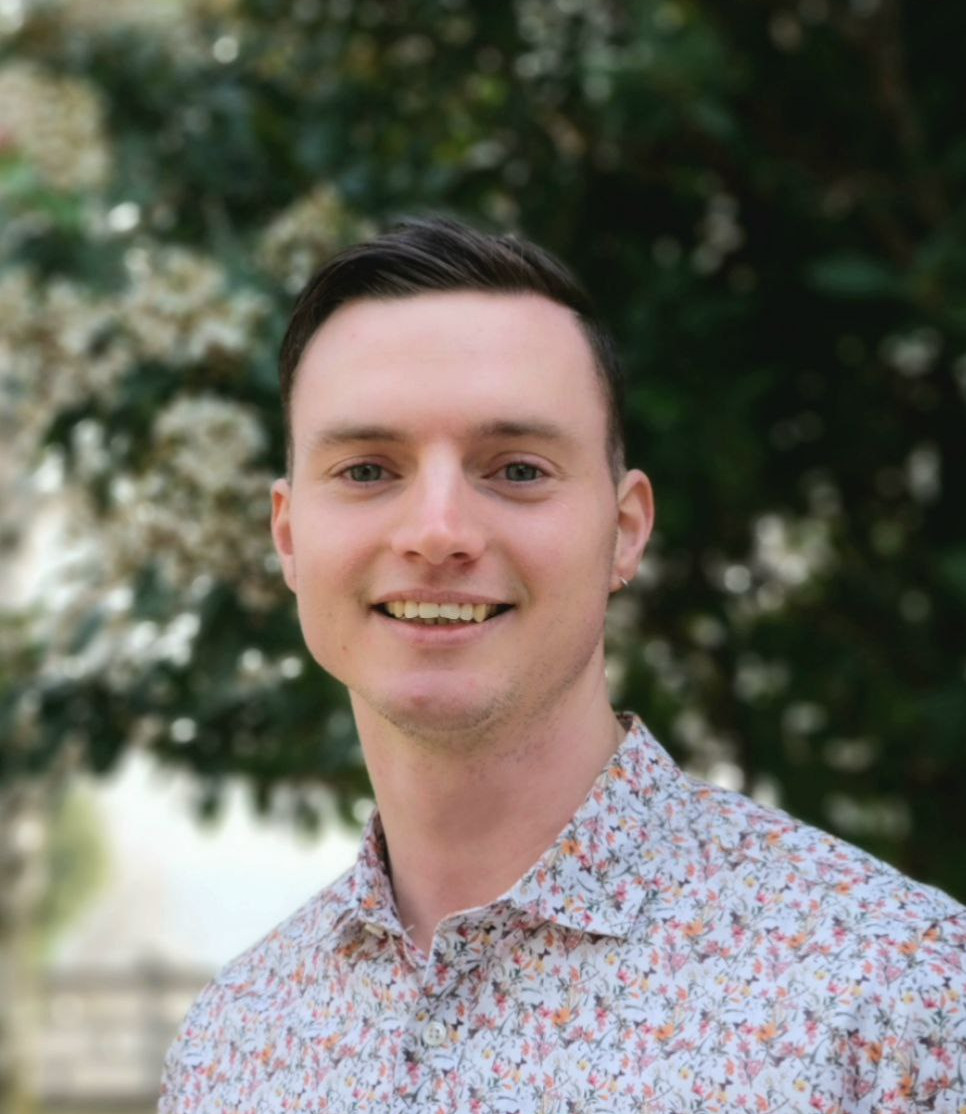}}]{Francesco Taioli} is a PhD Student of the National PhD programme in Artificial Intelligence at the Polytechnic of Turin in collaboration with the University of Verona, supervised by prof. Marco Cristani and prof. Alessandro Farinelli. He received
his MSc in Computer Engineering (cum laude) graduating from the University of Verona in 2022. His main research interests are in computer vision and deep learning, with a focus on improving the autonomy of intelligent agents.
\end{IEEEbiography}
\vspace*{-2\baselineskip}

\begin{IEEEbiography}[{\includegraphics[width=1in,height=1.2in,clip,keepaspectratio]{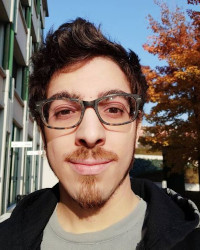}}]{Francesco Giuliari} (Student Member, IEEE)
is a PhD student at University of Genoa. He is currently affiliated with Istituto Italiano di Tecnologia under the supervision of Dr. Alessio Del Bue. He received his MSc in Computer Science from University of Verona in 2018. His main research interests are in computer vision, scene understanding and vision-based agent navigation.
\end{IEEEbiography}
\vspace*{-2\baselineskip}
\begin{IEEEbiography}[{\includegraphics[width=1in,height=1.25in,clip,keepaspectratio]{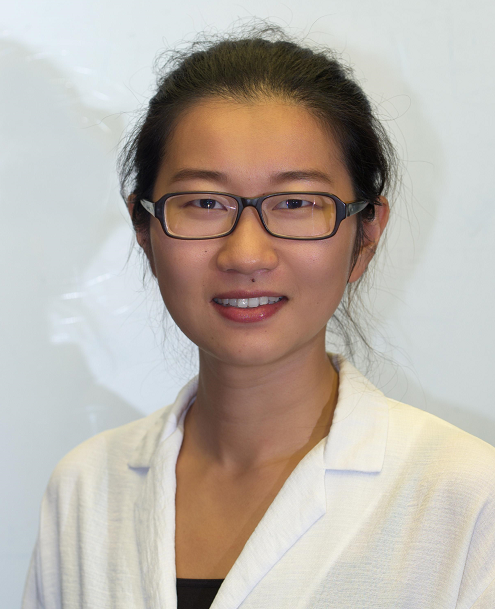}}]{Yiming Wang}
is a researcher in the Deep Visual Learning (DVL) unit at Fondazione Bruno Kessler (FBK). Her research mainly focuses on vision-based scene understanding that facilitates automation for social good. Yiming obtained her PhD in Electric Engineering from Queen Mary University of London (UK) in 2018, working on vision-based multi-agent navigation. Since 2018, she has worked as a post-doc researcher in the Pattern Analysis and Computer Vision (PAVIS) research line at Istituto Italiano di Tecnologia (IIT), working on topics related to active 3D vision. She has organised a couple of workshops on related topics and she is actively serving as reviewers for top-tier conferences and journals in both the Computer Vision and Robotics domains.
\end{IEEEbiography}
\vspace*{-2\baselineskip}

\begin{IEEEbiography}
[{\includegraphics[width=1in,height=1.2in,clip,keepaspectratio]{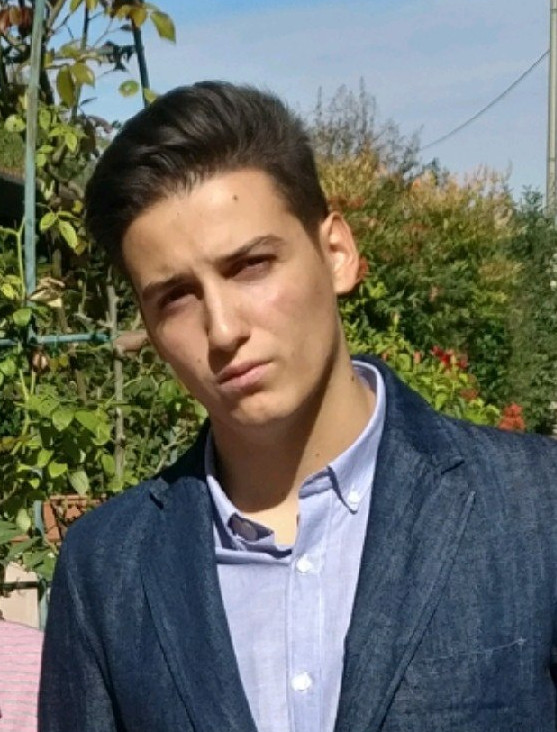}}]{Riccardo Berra} is a computer science graduate from the University of Verona, currently enrolled in the Master's degree program in Computer Engineering for Robotics and Smart Industry. Riccardo has collaborated several times in publications in the field of robotics and computer vision under the supervision of prof. Marco Cristani and prof. Francesco Setti.
\end{IEEEbiography}
\vspace*{-2\baselineskip}

\begin{IEEEbiography}[{\includegraphics[width=1in,height=1.2in,clip,keepaspectratio]{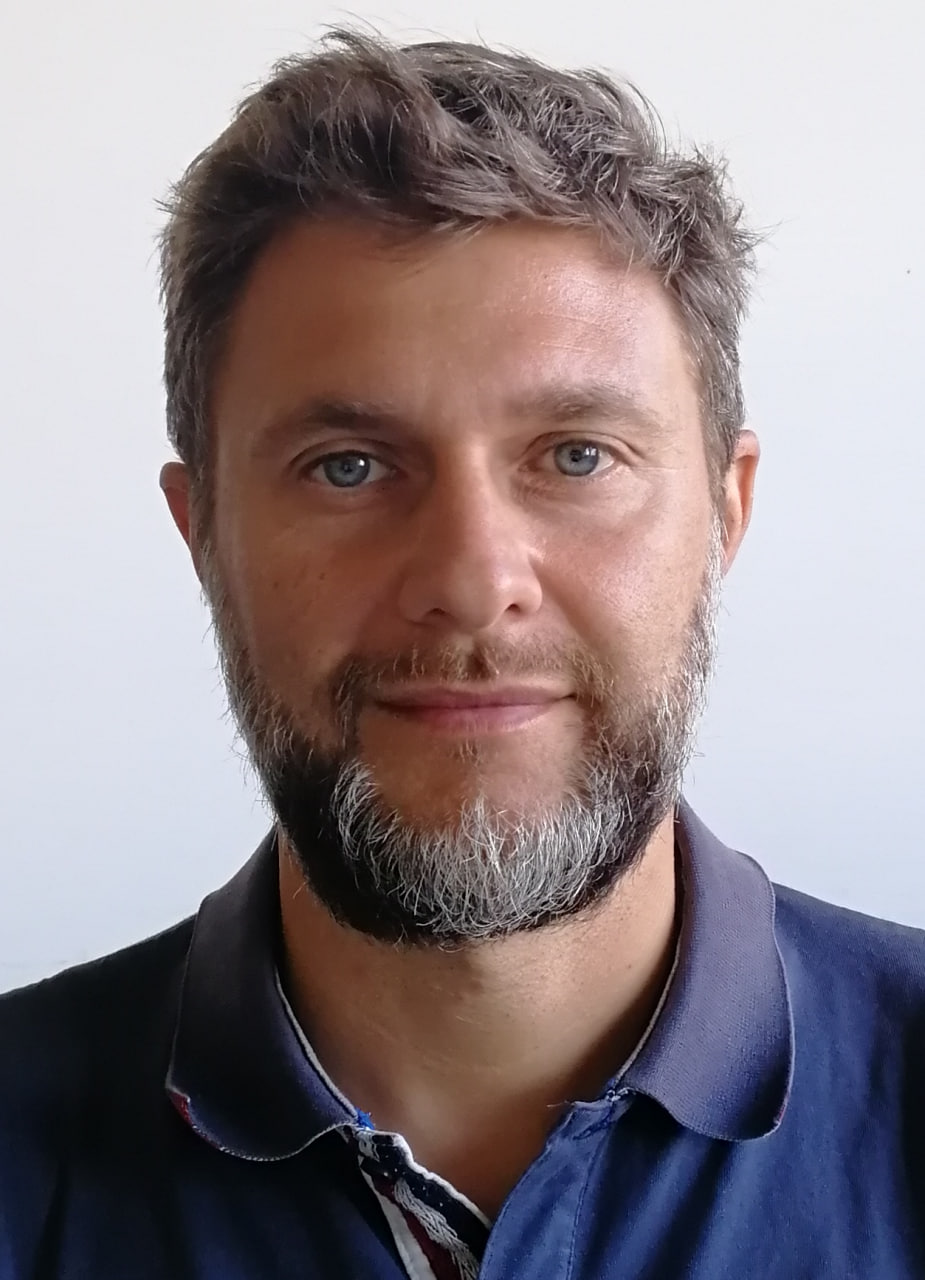}}]{Alberto Castellini} is assistant professor at the University of Verona, Dept. of Computer Science, since 2018. Before he was research fellow at Potsdam University/Max Planck Institute and University of Verona. His research interests include probabilistic planning under uncertainty, reinforcement learning, statistical learning and data analysis for intelligent systems (e.g., robotic/cyber-physical systems). He published in artificial intelligence journals and conferences (IEEE Int. Syst., Eng. App. of Artificial Intelligence, Rob. Aut. Systems, IJCAI, AAMAS, ICAPS, IROS).
\end{IEEEbiography}
\vspace*{-2\baselineskip}

\begin{IEEEbiography}[{\includegraphics[width=1in,height=1.25in,clip,keepaspectratio]{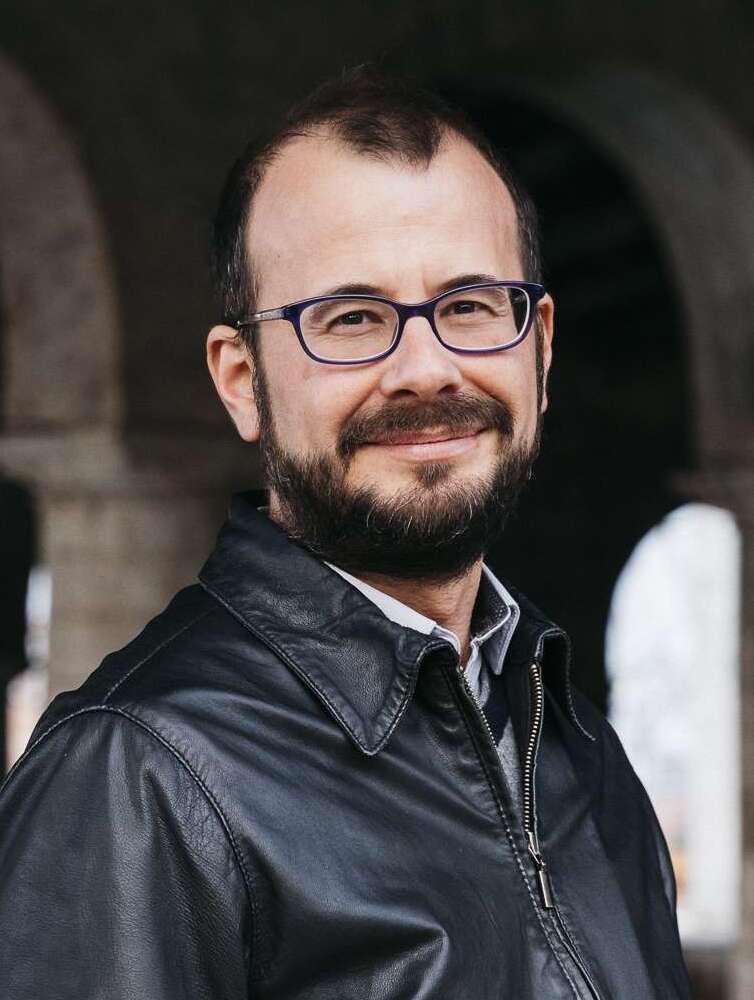}}]{Alessio Del Bue} (Member, IEEE) is a Tenured Senior Researcher leading the Pattern Analyisis and computer VISion (PAVIS) Research Line of the Italian Institute of Technology (IIT), Genoa, Italy. He is a coauthor of more than 100 scientific publications in refereed journals and international conferences on computer vision and machine learning topics. His current research interests include 3D scene understanding from multi-modal input (images, depth, and audio) 
to support the development of assistive artificial intelligence systems. He is a member of the technical committees of major computer vision conferences (CVPR, ICCV, ECCV, and BMVC). He serves as an Associate Editor for Pattern Recognition and Computer Vision and Image Understanding
journals. He is a member of ELLIS.
\end{IEEEbiography}
\vspace*{-2\baselineskip}

\begin{IEEEbiography}[{\includegraphics[width=1in,height=1.25in,clip,keepaspectratio]{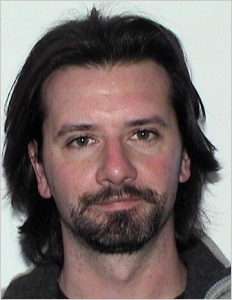}}]{Alessandro Farinelli} is a full professor at University of Verona, Department of Computer Science, since December 2019. His research interests focus on developing novel methodologies for Artificial Intelligence systems applied to robotics and cyber-physical systems. In particular, he focuses on multi-agent coordination, decentralised optimisation, reinforcement learning and data analysis for cyber-physical systems. Alessandro Farinelli was principal investigator for several national and international research projects in the broad area of Artificial Intelligence. 
\end{IEEEbiography}
\vspace*{-2\baselineskip}

\begin{IEEEbiography}[{\includegraphics[width=1in,height=1.1in,clip,keepaspectratio]{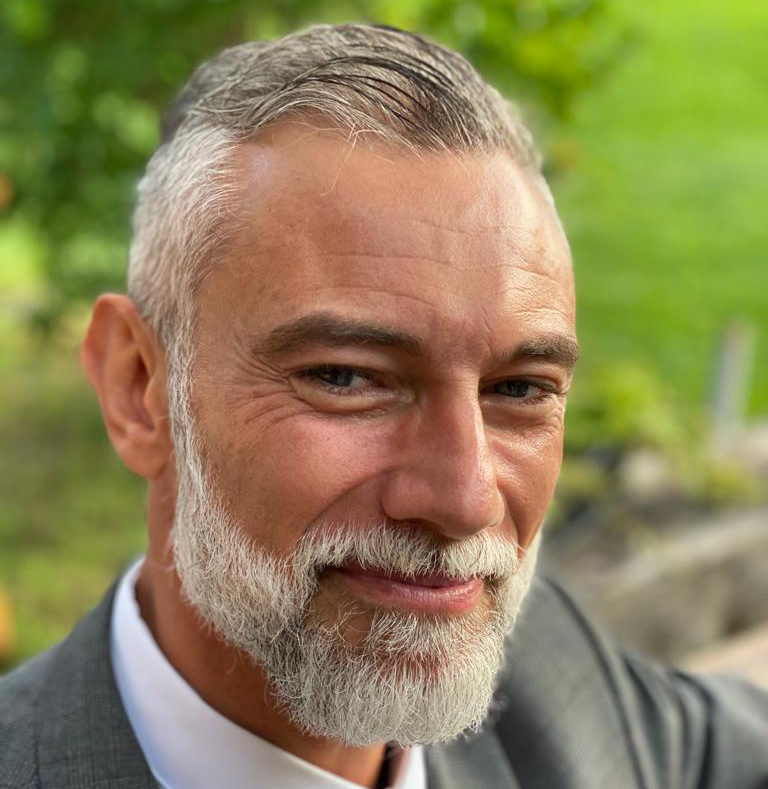}}]{Marco Cristani} is Full Professor (Professore Ordinario) at the Computer Science Department, University of Verona,
Associate Member at the National Research Council (CNR), External Collaborator at the Italian Institute of Technology (IIT). His main research interests are in statistical pattern recognition and computer vision, mainly in deep learning and generative modeling, with application to social signal processing and fashion modeling. On these topics he has published more than 200 papers. He has organised 11 international workshops, cofounded a spin-off company, Humatics, dealing with e-commerce for fashion. He is or has been Principal Investigator of several national and international projects, including PRIN and H2020 projects. He is IAPR fellow and member of IEEE.
\end{IEEEbiography}
\vspace*{-2\baselineskip}

\begin{IEEEbiography}[{\includegraphics[width=1in,height=1.2in,clip,keepaspectratio]{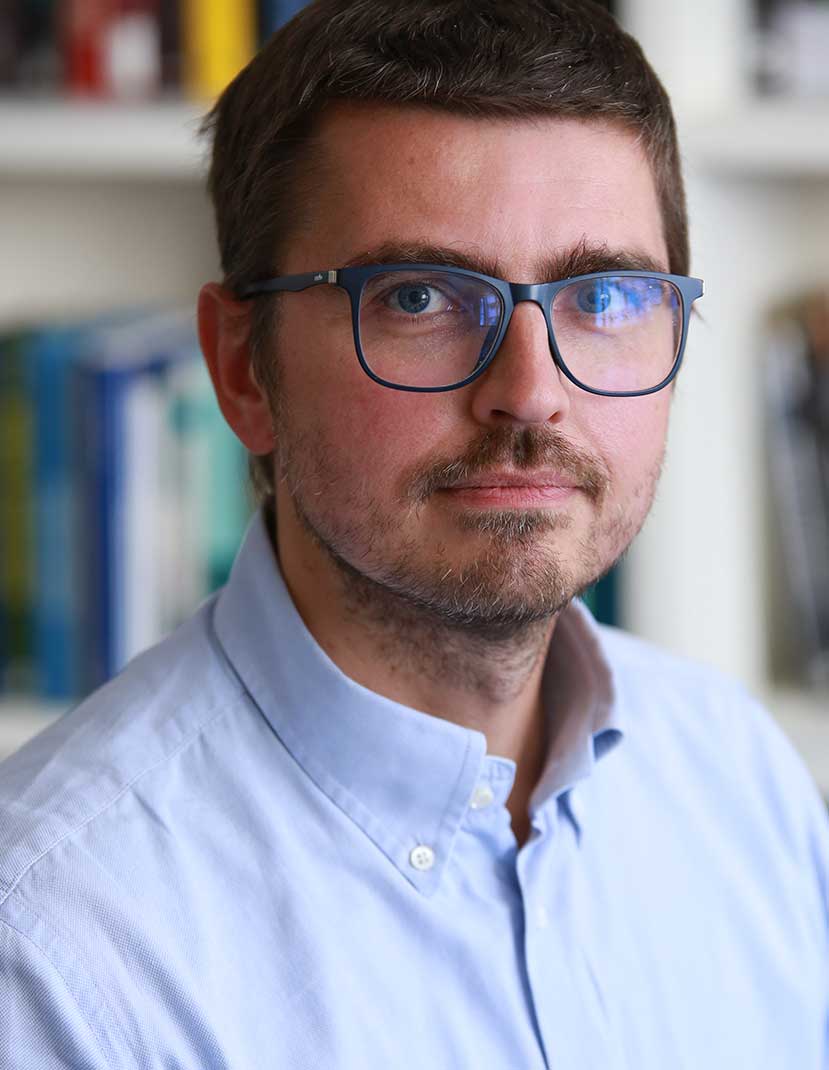}}]{Francesco Setti} is Assistant Professor at the University of Verona, Dept. of Computer Science and Associate Researcher of the Institute of Cognitive Science and Technology (ISTC-CNR).
His research interests include Computer Vision, Machine Learning, and Artificial Intelligence applied to Robotics an Manufacturing. He is a coauthor of more than 50 papers in international peer-reviewed journals and conferences. He is member of the IEEE, IAPR and BMVA, and he serves as reviewer for all the major machine learning conferences and journals (including CVPR, ICCV, ECCV, and TPAMI).
\end{IEEEbiography}
\vspace*{-2\baselineskip}

\end{document}